\documentclass[10pt,twocolumn,letterpaper]{article}

\usepackage[pagenumbers]{cvpr} 

\usepackage[dvipsnames]{xcolor}

\definecolor{cvprblue}{rgb}{0.21,0.49,0.74}
\usepackage[pagebackref,breaklinks,colorlinks,citecolor=cvprblue]{hyperref}
\newcommand{\myparagraph}[1]{\vspace{2pt}\noindent{\bf #1}}
\usepackage{indentfirst}
\usepackage{array,multirow,graphicx}
\usepackage{float}
\usepackage{amsmath}
\usepackage{mathtools}
\usepackage{bbm}
\usepackage{hhline}
\usepackage{boldline}
\usepackage{algorithm} 
\usepackage{algpseudocode} 
\usepackage{graphicx}
\usepackage{amssymb}

\usepackage{colortbl}
\usepackage{booktabs} 

\newcommand{\eagle}{EAGLE}
\newcommand{\eigenmap}{EiCue}

\newcommand*\samethanks[1][\value{footnote}]{\footnotemark[#1]}

\def\confName{CVPR}
\def\confYear{2024}

\title{EAGLE: Eigen Aggregation Learning for Object-Centric \\Unsupervised Semantic Segmentation}

\author{
Chanyoung Kim\thanks{Equal contribution}\quad Woojung Han\samethanks\quad Dayun Ju\quad Seong Jae Hwang\thanks{Corresponding author}\\ 
Yonsei University
\\{\tt\small \{chanyoung, dnwjddl, juda0707, seongjae\}@yonsei.ac.kr}
}

\begin{document}
\maketitle
\begin{abstract}

Semantic segmentation has innately relied on extensive pixel-level annotated data, leading to the emergence of unsupervised methodologies. Among them, leveraging self-supervised Vision Transformers for unsupervised semantic segmentation (USS) has been making steady progress with expressive deep features. Yet, for semantically segmenting images with complex objects, a predominant challenge remains: the lack of explicit object-level semantic encoding in patch-level features. This technical limitation often leads to inadequate segmentation of complex objects with diverse structures. To address this gap, we present a novel approach, \textbf{EAGLE}, which emphasizes object-centric representation learning for unsupervised semantic segmentation. Specifically, we introduce EiCue, a spectral technique providing semantic and structural cues through an eigenbasis derived from the semantic similarity matrix of deep image features and color affinity from an image. Further, by incorporating our object-centric contrastive loss with EiCue, we guide our model to learn object-level representations with intra- and inter-image object-feature consistency, thereby enhancing semantic accuracy. Extensive experiments on COCO-Stuff, Cityscapes, and Potsdam-3 datasets demonstrate the state-of-the-art USS results of EAGLE with accurate and consistent semantic segmentation across complex scenes.

\end{abstract}    
\vspace{-5pt}
\section{Introduction}
\label{sec:intro}
\vspace{-5pt}

\begin{figure}[t!]
  \centering
  \includegraphics[width = \linewidth]{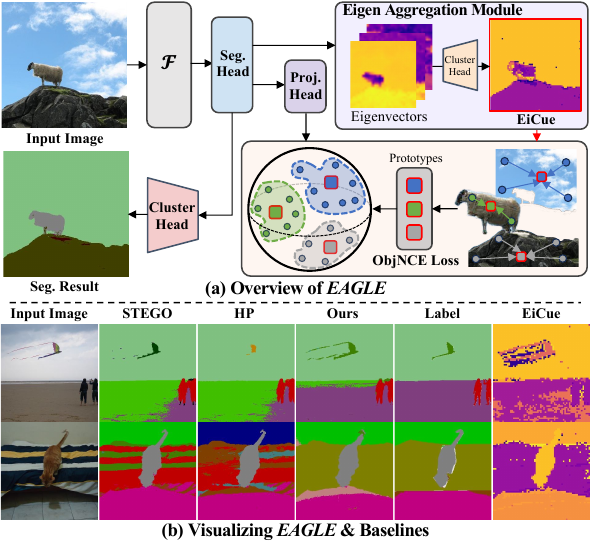}
\vspace{-22pt}
  \caption{
  We introduce \textbf{\textit{\eagle}}, \textit{\textbf{E}igen \textbf{AG}gregation \textbf{LE}arning} for object-centric unsupervised semantic segmentation. 
  \textbf{(a)} We first leverage the aggregated eigenvectors, named \eigenmap{}, to obtain the semantic structure knowledge of object segments in an image. Based on both semantic and structural cues from the \eigenmap{}, we compute object-centric contrastive loss to learn object-level semantic representation. \textbf{(b)} A visual comparison between \textit{\eagle{}} and other methods. Our object-level semantic segmentation results robustly identify objects with complex semantics (e.g., \texttt{blanket} with vivid stripe patterns) by exploiting strong semantic structure cues from \eigenmap{}.
  }
\vspace{-17pt}
  \label{fig:fig2_2a}
\end{figure}

Semantic segmentation plays a pivotal role in modern vision, fundamentally advancing an array of diverse areas including medical imaging~\cite{unet, jiang2018medical}, autonomous driving~\cite{autodrive1, teichmann2018multinet}, and remote sensing imagery~\cite{li2021multiattention, ding2020lanet}.
\let\thefootnote\relax\footnote{\scriptsize{Project Page:~\url{https://micv-yonsei.github.io/eagle2024/}}}
Nevertheless, its reliance on labeled data, while common across nearly all vision tasks, is especially problematic due to the laborious and time-consuming process of pixel-level annotation.
In response to this challenge, various studies in semantic segmentation tasks have drifted away from relying solely on human-labeled annotations by exploring weakly-supervised~\cite{vezhnevets2010towards, kervadec2019constrained, ahn2018learning, Beco, acr}, semi-supervised~\cite{alonso2021semi, ouali2020semi, lai2021semi}, and \textit{unsupervised semantic segmentation}~(USS) methodologies~\cite{iic, autoregressive, infoseg, picie, transfgu, vice, stego, HP}.

Among these learning schemes, the unsupervised approach of USS clearly stands as the most challenging case. 
Specifically, compared to the classical unsupervised segmentation methods (e.g., K-means clustering) which produce segments without explicit semantics, USS additionally aims to derive semantically consistent local features (e.g., patch-level features) that aid the further class assignment post-steps via clustering and the Hungarian matching algorithm. That is, semantically plausible local features result in accurate semantic segmentation results (e.g., Fig.~\ref{fig:fig2_2a}b), but in USS, this must be achieved \textit{without any labels}.

Despite the glaring challenge, steady progress has been shown in USS. For example, initial pioneering works have emerged to maximize the mutual information across the two different views of a single image~\cite{iic, autoregressive}.
Recently, network-based techniques such as STEGO~\cite{stego} have focused on deriving patch-level semantic features with a self-supervised pretrained model~\cite{dino}, showing a significant improvement compared to previous methods~\cite{iic, picie, transfgu}.
However, while these methodologies have advanced USS, unresolved shortcomings still remain.

In particular, the recent network-based methods often leverage a self-supervised Vision Transformer (ViT) to learn patch-level features. While their patch-level features proved to be useful for further USS inference steps (e.g., K-means), the underlying object-level semantics are not explicitly imposed in these patch-level features. To grasp the ``object-level semantics'', consider an example of a \texttt{blanket} 
object as shown in Fig.~\ref{fig:fig2_2a}b second row. As with any object, \texttt{blanket} may easily appear with varying colors and textures across different images. Without proper object-level semantics, features corresponding to varying regions of \texttt{blanket} may result in vastly different feature representations. Ideally, though, the features corresponding to all kinds of \texttt{blanket} should be mapped to similar features, namely, object-level semantics. Thus, without carefully imposed object-level semantics, complex objects with diverse structures and shapes may easily be partitioned into multiple segments with wrong class labels or be merged with nearby segments of different class labels. Thus, in USS, an immense effort must be paid to learn the local features (e.g., patch-level) with strong object-level semantics.

Our object-centric representation learning for USS aims to capture such object-level semantics.
Specifically, we first need a semantic or structure cue in the object-centric view.
Several previous works utilized clustering methods such as K-means or superpixel to obtain semantic cues~\cite{segsort}, however, they mainly fixated on the generic image patterns, not the object's semantic or structural representation.
Here, we propose \eigenmap{} which provides semantic and structural cues of objects via eigenbasis.
Specifically, we utilize the semantic similarity matrix obtained from the projected deep image features obtained from ViT~\cite{vit, dino} and the color affinity matrix of the image to construct the graph Laplacian. The corresponding eigenbasis captures the underlying \textit{semantic structures} of objects~\cite{zhang2023hivit, swint}, providing soft guidance to the subsequent object-level feature refinement step.

Recall that accurate object-level semantics of an object must be consistent across images. Our object-centric contrastive learning framework explicitly imposes these traits with a novel object-level contrastive loss. Specifically, based on the object cues from \eigenmap{}, we derive learnable prototypes for each object which enables intra- and inter-image object-feature consistency. 
Through this comprehensive learning process, our model effectively captures the inherent structures within images, allowing it to precisely identify semantically plausible object representations, the key to advancing modern feature-based USS. 

\vspace{3pt}
\noindent\textbf{Contributions.} Our main contributions are as follows:
\begin{itemize}

\item{We propose \eigenmap{}, using a learnable graph Laplacian, to acquire a more profound understanding of the underlying semantics and structural details within images.}
\item{We design an object-centric contrastive learning framework that capitalizes on the spectral basis of \eigenmap{} to construct robust object-level feature representations.}
\item{We demonstrate that our \textit{\eagle{}} achieves state-of-the-art performance on unsupervised semantic segmentation, supported by a series of comprehensive experiments. }
\end{itemize}

\vspace{-3pt}
\section{Related Work}
\label{sec:formatting}

\subsection{Unsupervised Semantic Segmentation}
\vspace{-5pt}
\begin{figure*}[t!]
    \begin{center}
        \includegraphics[width=\textwidth]{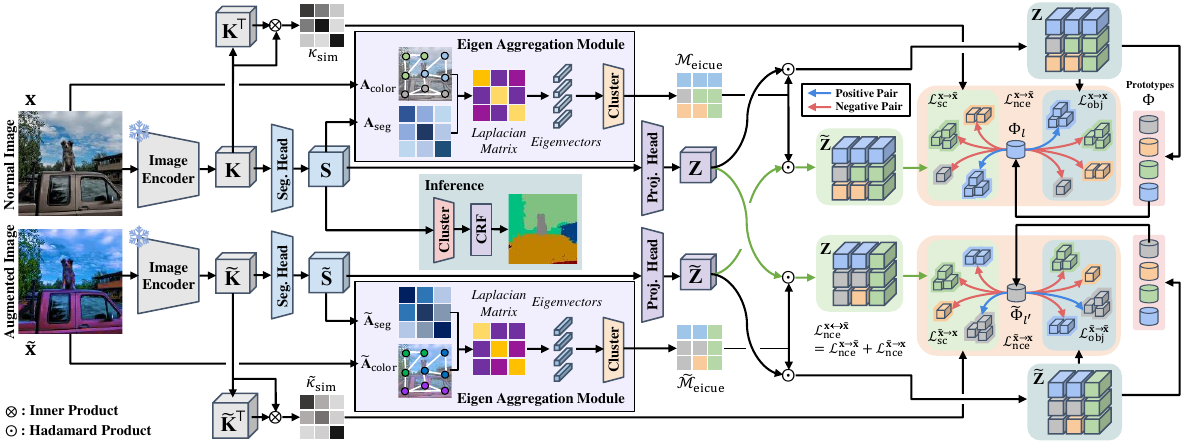}
    \end{center}
    \vspace{-17pt}
    \caption{The pipeline of \textit{\textbf{\eagle{}}}. Leveraging the Laplacian matrix, which integrates hierarchically projected image key features and color affinity, the model exploits eigenvector clustering to capture object-level perspective cues defined as $\mathcal{M}_{\text{eicue}}$ and $\Tilde{\mathcal{M}}_{\text{eicue}}$. Distilling knowledge from $\mathcal{M}_{\text{eicue}}$, our model further adopts an object-centric contrastive loss, utilizing the projected feature $\mathbf{Z}$ and $\Tilde{\mathbf{Z}}$. The learnable prototype $\Phi$ assigned from $\mathbf{Z}$ and $\Tilde{\mathbf{Z}}$, acts as a singular anchor that contrasts positive objects and negative objects. Our object-centric contrastive loss is computed in two distinct manners: intra($\mathcal{L}_{\text{obj}}$)- and inter($\mathcal{L}_{\text{sc}}$)-image to ensure semantic consistency.
    }
    \label{fig:overview}
    \vspace{-15pt}
\end{figure*}

Semantic segmentation plays a crucial role in vision by assigning distinct class labels to pixels. 
Yet, while the segmentation performance strongly correlates with the label quality, acquiring precise pixel-level ground truth labels is a challenge on its own, especially for images with complex structures. This naturally led to numerous attempts to perform semantic segmentation in an unsupervised manner~\cite{iic, autoregressive, infoseg, picie, transfgu, vice, stego, HP}, that is, with no labels.
For instance, early works such as IIC~\cite{iic} and AC~\cite{autoregressive} utilized mutual information, while subsequent approaches like InfoSeg~\cite{infoseg} and PiCIE~\cite{picie} integrated diverse features for enhanced pixel learning.
Recent studies have adopted self-supervised, pretrained ViT models like DINO~\cite{dino} for top-down feature extraction.
Namely, STEGO~\cite{stego} demonstrated a major step forward by distilling unsupervised features into discrete semantic labels with the DINO backbone.
HP~\cite{HP} interestingly utilizes contrastive learning to enhance semantic correlations among patch-level regions, but this patch-level~(local) refinement holds little object-level understanding.

\subsection{Spectral Techniques for Segmentation}
\vspace{-5pt}
Predating the aforementioned methods for semantic segmentation, spectral techniques have long been offering insights into diverse segmentation challenges in vision. Spanning some early pioneering works~\cite{shi2000normalized, ng2001spectral, pappas1989adaptive, luo2003spatial} to contemporary efforts~\cite{dhanachandra2015image, koohpayegani2021mean, barbato2022unsupervised, shen2016real, manavalan2011trus}, these techniques share a common aim: to exploit the intrinsic spectral signatures embedded within image regions.
These graph-theoretic approaches are methodologically influenced by the affinity matrix quality, which gave rise to recent methods utilizing the network features from the pretrained deep models.
For instance, Deep Spectral Methods~\cite{melas2022deep} builds powerful Laplacian eigenvectors from the feature affinity matrix, while EigenFunction~\cite{EigenFunction} exploits the network-based learnable eigenfunctions to produce spectral embeddings.
Despite steadily discovering the effectiveness of spectral methods on deep features for capturing complex object structures, their object-level semantics still require additional methodological efforts, e.g., contrastive learning.

\vspace{-5pt}
\subsection{Object-centric Contrastive Learning}
\vspace{-7pt}
Contrastive learning approaches aim to maximize feature similarities between similar units while minimizing them between dissimilar ones. 
In the task of semantic segmentation, patch-level representation learning~\cite{VADeR, denseCL, PixPro} is widely used. 
However, this approach tends to overemphasize fine details while neglecting high-level concepts~(i.e., semantic relations between objects). 
This leads object-level contrastive learning methods~\cite{maskcontrast, zadaianchuk2023unsupervised, cast, DetCon, hierarchicalGrouping, wen2022slotcon, seitzer2023bridging} to focus on balancing detailed perception with an object-centric view, identifying objects in an unsupervised manner.
For instance, MaskContrast~\cite{maskcontrast} and COMUS~\cite{zadaianchuk2023unsupervised} use unsupervised saliency to make pixel embeddings, while Odin~\cite{Odin} and DetCon~\cite{DetCon} utilize K-means clustering and heuristic masks for sample generation, respectively.
Refining this, SlotCon~\cite{wen2022slotcon} assigned pixels to learn slots for semantic representation, and DINOSAUR~\cite{seitzer2023bridging} further improved it by reconstructing self-supervised pretrained features in the decoder, instead of the original inputs.
However, these methods~\cite{wen2022slotcon, seitzer2023bridging} rely solely on slots, potentially overlooking high-level image features.
In contrast, our approach distills knowledge from clustered eigenvectors derived from a similarity matrix-based Laplacian capturing their object semantic relationships.

\vspace{-5pt}
\section{Methods}
\vspace{-7pt}
As we begin describing our full pipeline shown in Fig.~\ref{fig:overview}, let us first cover the core USS framework based on pretrained models as in prior works~\cite{stego, HP}.
\vspace{-3pt}
\subsection{Preliminary}
\label{sec:prelim}
\vspace{-5pt}

\noindent\textbf{Unlabeled Images.} Our approach is built exclusively upon a set of images, \textit{without} any annotations, denoted as $\mathbf{X}=\{\mathbf{x}_b\}_{b=1}^B$, where $B$ is the number of training images within a mini-batch.
We also utilize a photometric augmentation strategy $P$ to obtain an augmented image set $\Tilde{\mathbf{X}} = \{\mathbf{\Tilde{x}}_b\}_{b=1}^B = P(\mathbf{X)}$.

\myparagraph{Pretrained Features $\mathbf{K}$.} Then, for each input image $\mathbf{x}_b$, we use a self-supervised pretrained vision transformer~\cite{dino} as an image encoder $\mathcal{F}$ to obtain hierarchical attention key features from the last three blocks as $\mathbf{K}_{L-2} = \mathcal{F}_{L-2}(\mathbf{x}_b)$, $\mathbf{K}_{L-1} = \mathcal{F}_{L-1}(\mathbf{x}_b)$, $\mathbf{K}_{L} = \mathcal{F}_{L}(\mathbf{x}_b)$, where $L-2$, $L-1$, $L$ is the third-to-last layer, the second-to-last layer, and the last layer, respectively.
Then, we concatenate them into a single attention tensor $\mathbf{K}=[\mathbf{K}_{L-2}; \mathbf{K}_{L-1}; \mathbf{K}_{L}] \in \mathbb{R}^{H\times W\times D_K}$.
Similarly, we apply the same procedure for the augmented image $\Tilde{\mathbf{x}}$ and obtain its attention tensor $\Tilde{\mathbf{K}}\in \mathbb{R}^{H\times W\times D_K}$.

\myparagraph{Semantic Features $\mathbf{S}$.} Although $\mathbf{K}$ contains some structural information about the objects based on the attention mechanism, this is known for insufficient semantic information to be considered for direct inference. Thus, for further feature refinement, we compute the semantic features $\mathbf{S} = \mathcal{S}_\theta(\mathbf{K})\in \mathbb{R}^{H\times W\times D_S}$ and $\Tilde{\mathbf{S}} = \mathcal{S}_\theta(\Tilde{\mathbf{K}})\in \mathbb{R}^{H\times W\times D_S}$, where $\mathcal{S}_\theta:\mathbb{R}^{H\times W\times D_K}\rightarrow\mathbb{R}^{H\times W\times D_S}$ is a learnable nonlinear segmentation head.
For brevity, the total number of patches, denoted as $H \times W$, will be referred to as $N$.

\myparagraph{Inference.} During the inference time, given a new image, its semantic feature $\mathbf{S}$ becomes the basis of further clustering for the final semantic segmentation output with conventional evaluation setups such as the K-means clustering and linear probing.
Thus, as with prior pretrained feature-based USS works~\cite{stego, HP}, training $\mathcal{S}_\theta$ to output strong semantic features $\mathbf{S}$ in an unsupervised manner is the basic framework of contemporary USS frameworks. 
We next describe the remainder of the pipeline in Fig.~\ref{fig:overview} which corresponds to our methodological contributions for producing powerful \textit{object-level} semantic features.

\vspace{-3pt}
\subsection{\eigenmap{} via the Eigen Aggregation Module}
\label{sec:eigenclusteringmodule}
\vspace{-5pt}
Intuition tells us that the ``semantically plausible'' object-level segments are groups of pixels precisely capturing the object structure, even under complex structural variance. For instance, a \texttt{car} segment must contain all of its parts including the windshield, doors, wheels, etc. which may all appear in different shapes and views. However, without pixel-level annotations that provide object-level semantics, this becomes an extremely challenging task of inferring the underlying structure with zero object-level structural prior. 

\begin{figure}[t!]
  \centering
  \includegraphics[width = \linewidth]{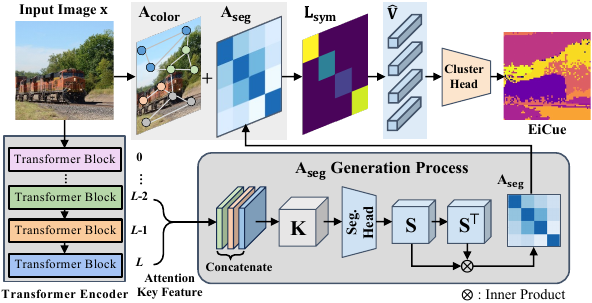}
\vspace{-20pt}
  \caption{An illustration of the \eigenmap{} generation process. From the input image, both color affinity matrix $\mathbf{A}_\text{color}$ and semantic similarity matrix $\mathbf{A}_\text{seg}$ are derived, which are combined to form the Laplacian $\mathbf{L}_\text{sym}$. 
  An eigenvector subset $\hat{\mathbf{V}}$ of $\mathbf{L}_\text{sym}$ are clustered to produce \eigenmap{}.
  } 
\vspace{-17pt}
  \label{fig:eigenmodule}
\end{figure}

\begin{figure}[t!]
  \centering
  \includegraphics[width = \linewidth]{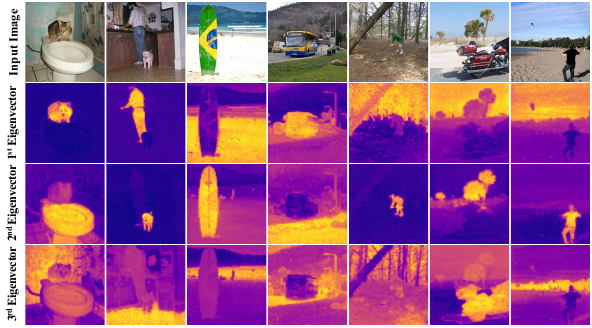}
  \vspace{-20pt}
  \caption{Visualizing eigenvectors derived from $\mathbf{S}$ in the Eigen Aggregation Module. These eigenvectors not only distinguish different objects but also identify semantically related areas, highlighting how \eigenmap{} captures object semantics and boundaries effectively.
  }
  \vspace{-15pt}
  \label{fig:eigenvector}
\end{figure}

From this realization, our model \textit{\eagle{}} first aims to derive a strong yet simple semantic structural cue, namely, \eigenmap{}, based on the eigenbasis of the feature similarity matrix as illustrated in Fig.~\ref{fig:eigenmodule}. Specifically, we use the well-known Spectral Clustering~\cite{cheeger2015lower, shi2000normalized, ng2001spectral} to obtain unsupervised feature representations that capture the underlying non-linear structures for handling data with complex patterns. This classically operates only in the color space but may easily extend to utilize the similarity matrix constructed from any features. We observed that such a spectral method becomes especially useful for complex real-world images as in Fig.~\ref{fig:eigenvector}.

\myparagraph{EiCue Construction.} Let us describe the process of constructing \eigenmap{} in detail as shown in Fig.~\ref{fig:eigenmodule}. 
The overall framework generally follows the vanilla spectral clustering: (1) from an adjacency matrix $\mathbf{A}$, (2) construct the graph Laplacian  $\mathbf{L}$, and (3) perform the eigendecomposition on $\mathbf{L}$ to derive the eigenbasis $\mathbf{V}$ from which the eigenfeatures are used for the clustering. We describe each step below.

\vspace{-7pt}
\subsubsection{Adjacency Matrix Construction}
\vspace{-8pt}
Our adjacency matrix consists of two components: (1) color affinity matrix and (2) semantic similarity matrix.

\vspace{3pt}
\noindent\textbf{(I) Color Affinity Matrix $\mathbf{A}_{\text{color}}$}: The color affinity matrix leverages the RGB values of the image $\mathbf{x}$. 
    The color affinity matrix is computed by the color distance. It utilizes the Euclidean distance between patches, where $p$ and $q$ are specific patch positions within the image. 
    Here, $\ddot{\mathbf{x}}\in\mathbb{R}^{H\times W\times 3}$ denotes a resized version of $\mathbf{x}$, scaled from its original image resolution to patch resolution, to ensure compatibility with the dimensions of other adjacency matrices.
    The resulting color affinity matrix, $\mathbf{A}_{\text{color}}\in\mathbb{R}^{N\times N}$ thus captures the pairwise relationship between the patches based on the colors. 
    Specifically, we use the RBF kernel as the distance function $\mathbf{A}_{\text{color}}(p,q) = \text{exp} \left(-{\left\| \ddot{\mathbf{x}}(p) - \ddot{\mathbf{x}}(q)\right\|}_2 / { 2{\sigma_c}^2} \right)$
    where $\sigma_c>0$ is a free hyperparameter.
    Further, to ensure that only nearby patches influence each other's affinity values, we hard-constrain the maximum distance of the patch pairs such that we only compute the affinity between the patch pairs with a predefined spatial distance.
     
    \myparagraph{(II) Semantic Similarity Matrix $\mathbf{A}_{\text{seg}}$}: 
    The semantic similarity matrix, denoted as $\mathbf{A}_{\text{seg}}\in\mathbb{R}^{N\times N}$, is formed by the product of tensor $\mathbf{S}$ and its transpose $\mathbf{S}^\top$. 
    Tensor $\mathbf{S}$ is derived by hierarchically concatenating key attention features from the last three layers of a pretrained vision transformer, as processed through the segmentation head $\mathcal{S}_\theta$.
    
    \myparagraph{(III) Adjacency Matrix $\mathbf{A}$}: The final adjacency matrix $\mathbf{A}$ is the sum of $\mathbf{A}_{\text{color}}$ and $\mathbf{A}_{\text{seg}}$: $\mathbf{A}=\mathbf{A}_{\text{color}}+\mathbf{A}_{\text{seg}}$, which is also applicable to $\Tilde{\mathbf{A}}$.
    Our adjacency matrix amalgamates the high-level color information and the network-based deep features to characterize semantic-wise relations. 
    The use of the image-based $\mathbf{A}_\text{color}$ preserves the image's structural integrity and also complements the contextual information of the image. Following this, the incorporation of the learnable tensor $\textbf{S}$ for the $\mathbf{A}_\text{seg}$ further strengthens this aspect, enhancing the semantic interpretation of the object without compromising the structural integrity and serving as a vital cue for our learning process.

\vspace{-7pt}
\subsubsection{Eigendecomposition}
\vspace{-8pt}
To construct \eigenmap{} based on $\mathbf{A}$, a Laplacian matrix is created. Formally, the Laplacian Matrix is expressed as $\mathbf{L}=\mathbf{D}-\mathbf{A}$, where $\mathbf{D}$ is the degree matrix of $\mathbf{A}$ defined as $\mathbf{D}(i,i) = \sum_{j=1}^{N} \mathbf{A}(i,j)$.
In our approach, we utilize the normalized Laplacian matrix for its enhanced clustering capabilities.
The symmetric normalized Laplacian matrix $\mathbf{L}_{\text{sym}}$ are defined as $\mathbf{L}_{\text{sym}} = \mathbf{D}^{-\frac{1}{2}}\mathbf{L}\mathbf{D}^{-\frac{1}{2}}$. Then, via eigendecomposition on $\mathbf{L}_{\text{sym}}$, the eigenbasis $\mathbf{V}\in\mathbb{R}^{N\times N}$ is computed, where each column corresponds to a unique eigenvector. We then extract $k$ eigenvectors corresponding to the $k$ smallest eigenvalues and concatenate them into $\mathbf{\hat{V}}\in\mathbb{R}^{N\times k}$ where the $i^{th}$ row corresponds to the $k$ dimensional eigenfeature of the $i^{th}$ patch.

\vspace{-5pt}
\subsubsection{Differentiable Eigen Clustering}
\vspace{-7pt}
After obtaining eigenvectors $\mathbf{\hat{V}}$, we perform the eigenvectors clustering process and extract the \eigenmap{} denoted as  $\mathcal{M}_{\text{eicue}}\in\mathbb{R}^{N}$. 
To cluster eigenvectors, we leverage a minibatch K-means algorithm based on cosine distance~\cite{clusteringprocess} between $\mathbf{\hat{V}}$ and $\mathbf{C}$, denoted as $\mathbf{P} = \mathbf{\hat{V}}\mathbf{C}$. 
Centers of clusters $\mathbf{C}\in\mathbb{R}^{k\times C}$ are composed of learnable parameters.
To learn $\mathbf{C}$, we further trained with a loss defined as follows: 

\begin{equation}\small
\label{eq_eigenloss}
    \mathcal{L}_{\text{eig}}^{\mathbf{x}} = -\frac{1}{N} \sum_{i=1}^{N} \bigg( \sum_{c=1}^C \Psi_{ic} \mathbf{P}_{ic} \bigg),
\end{equation}
where $C$ denotes pre-defined number of classes, $\Psi := \textit{softmax}(\mathbf{P})$ and $\mathbf{P}_{ic}$ and $\Psi_{ic}$ represents the $i^{th}$ patch and the $c^{th}$ cluster number of $\mathbf{P}$ and $\Psi$. 
We apply same procedure to augmented image $\mathbf{\Tilde{x}}$ to get $\mathcal{L}_{\text{eig}}^{\mathbf{\Tilde{x}}}$.
By minimizing $\mathcal{L}_{\text{eig}} = \frac{1}{2}(\mathcal{L}_{\text{eig}}^{\mathbf{x}} + \mathcal{L}_{\text{eig}}^{\mathbf{\Tilde{x}}})$, we can obtain centers of clusters that enable more effective clustering.
Then we obtain \eigenmap{} as
\begin{equation}\small
    \mathcal{M}_{\text{eicue}}(i) = \underset{c}{\text{argmax}} \bigg(\mathbf{P}_{ic} - \log \bigg( \sum_{c'=1}^C \exp(\mathbf{P}_{ic'}) \bigg) \bigg).
\end{equation}

As the precision of cluster centroids improves, \eigenmap{} facilitates the mapping of patch $i$ to its corresponding object based on semantic structure. 
This serves as a meaningful cue to stress semantic distinctions between different objects, thereby enhancing the discriminative power of the feature embeddings.

\myparagraph{Remark.} 
While similar to previous work~\cite{melas2022deep} in using eigendecomposition, our approach differs by enhancing feature vectors $\mathbf{S}$ with a trainable segmentation head, unlike their reliance on static vectors (i.e., $\mathbf{K}$). 
Our method enhances $\mathbf{S}$ learnable and adaptable via differentiable eigen clustering, allowing the graph Laplacian and object semantics to evolve. This dynamic integration of \eigenmap{} into the learning process distinctly separates our methodology from prior applications.

\vspace{-3pt}
\subsection{\eigenmap{}-based ObjNCELoss }\label{sec:ObjNCELoss}
\vspace{-5pt}

\begin{figure*}[t!]
    \begin{center}
        \includegraphics[width=\textwidth]{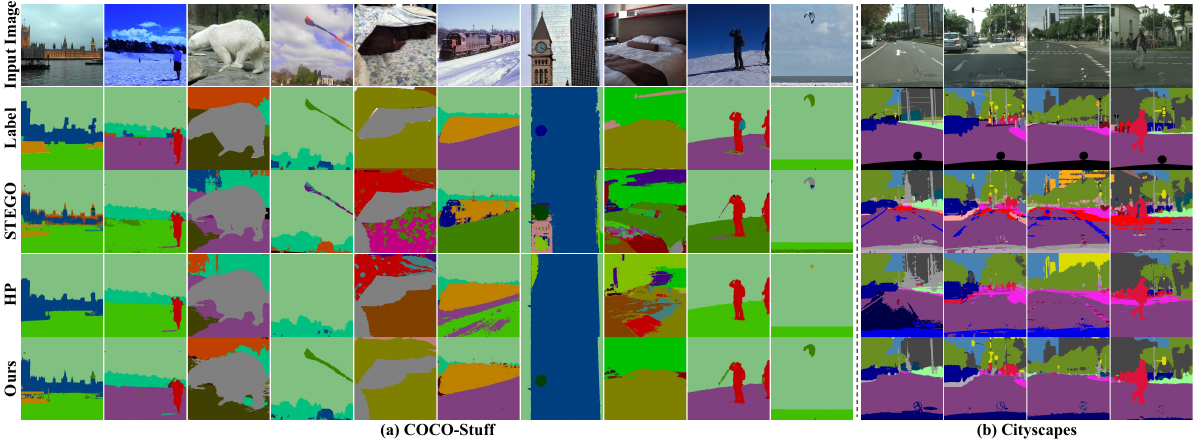}
    \end{center}
    \vspace{-20pt}
    \caption{
    A qualitative comparison of the (a) COCO-Stuff~\cite{coco} and (b) Cityscapes~\cite{cityscapes} datasets trained using ViT-S/8 and ViT-B/8 as a backbone, respectively. 
    The comparison included previous state-of-the-art USS approaches, STEGO~\cite{stego}, HP~\cite{HP}, and ours.
    }
    \label{fig:coco}
    \vspace{-15pt}
\end{figure*}

For a successful semantic segmentation task, it is important not only to classify the class of each pixel accurately but also to aggregate object representation
and create a segmentation map that reflects object semantic representations.
From this perspective, learning relationships in an object-centric view is especially crucial in semantic segmentation tasks.
To capture the complex relationships between objects, our approach incorporates an object-centric contrastive learning strategy, named \textit{\textbf{ObjNCELoss}}, guided by \eigenmap{}. 
This strategy is designed to refine the discriminative capabilities of feature embeddings $\mathbf{S}$, emphasizing the distinctions among various object semantics.
Before proceeding, we map both the projected feature $\mathbf{Z}\in\mathbb{R}^{N\times D_Z}$ and $\Tilde{\mathbf{Z}}\in\mathbb{R}^{N\times D_Z}$, using the linear projection head $\mathcal{Z}_\xi$, derived from the reshaped $\mathbf{S}\in\mathbb{R}^{N\times D_S}$ and $\Tilde{\mathbf{S}}\in\mathbb{R}^{N\times D_S}$, respectively. 
While the actual dimension sizes of $D_S$ and $D_Z$ are kept the same, we use different notations for ease of explanation.

\vspace{-10pt}
\subsubsection{Object-wise Prototypes}
\vspace{-7pt}
To extract the representative object level semantic features from projected feature $\mathbf{Z}$, we construct adaptable prototypes $\Phi_l$ based on the object $l$ in aforementioned \eigenmap{}. As we describe next, semantically representative prototypes become the anchors for either pulling objects with similar semantics while pushing away the different ones.

Let us describe how $\Phi$ is derived, which represents object semantics from $\mathbf{Z}$.
We first update the object-wise prototypes through the projected feature $\mathbf{Z}$ and a given $\mathcal{M}_{\text{eicue}}$, derived from the clustered eigenbasis.
Formally, for each object $l$ obtained from $\mathcal{M}_{\text{eicue}}$, the mask $M_l$ is defined as $M_l(i) = 1 \text{ if } \mathcal{M_{\text{eicue}}}(i) = l, \text{ and } 0 \text{ otherwise}$, 
where $i$ represents each position in $\mathcal{M}_{\text{eicue}}$.
Then, applying the mask $M_l$ to the projected feature tensor $\mathbf{Z}$ gives $\mathbf{Z}_l = \mathbf{Z} \odot M_l$, where $\odot$ denotes the Hadamard product and $\mathbf{Z}_l$ represents a collection of feature representations from $\mathbf{Z}$ corresponding to object $l$.
Next, we compute medoid to select a single vector from $\mathbf{Z}_l$, which then becomes the prototype $\Phi_l$.
Let $\mathcal{I}_l$ be the set of indices where $M_l^{(i\in \mathcal{I}_l)}=1$ to only consider the indices of object $l$. $\mathbf{Z}_l^{(i)}$ indicates the $i$-th feature vector of $\mathbf{Z}_l$.
Then, the prototype $\Phi_l$ from the masked tensor $\mathbf{Z}_l$ is
\begin{equation}\small
\small\Phi_l = \mathbf{Z}_l^{(m^*)} \ \text{ for \ } m^*=\underset{m \in \mathcal{I}_l}{\mathrm{argmin}} \sum_{i \in \mathcal{I}_l} \big\| \mathbf{Z}_l^{(m)} - \mathbf{Z}_l^{(i)} \big\|_2.
\end{equation}
Thus, $\Phi_l$ acts as the semantic vector of object $l$, serving as an anchor for the following object-centric contrastive loss.

\vspace{-5pt}
\subsubsection{Object-centric Contrastive Loss}
\vspace{-7pt}
Once we compute prototypes, we then step towards object-centric contrastive loss between prototypes $\Phi$ and feature vectors $\mathbf{Z}$. 
Specifically, we compute object-centric contrastive loss defined as follows:
\begin{equation}\small
\label{eq:l_obj}
\small \mathcal{L}_{\text{obj}}^{\mathbf{x}\rightarrow {\mathbf{x}}} = \frac{1}{N} \sum_{i=1}^{N} w_{\text{obj}}^{(i)} \left[ -\log \left( \frac{\exp((\mathbf{Z}_l^{(i)} \cdot \Phi_{l})/ \tau )}{\sum_{\substack{j=1, j\ne l}}^C \exp((\mathbf{Z}_j \cdot \Phi_l )/ \tau)} \right) \right],
\end{equation}
where $C$ is the total number of unique predicted objects in $\mathcal{M}_{\text{eicue}}$, $(\cdot)$ denotes the cosine similarity, and $\tau>0$ is the temperature scalar. 
To emphasize the influence of feature vectors with high similarity and direct the model's focus toward them, we weigh the loss based on the similarity information between vectors.
The weight $w_{\text{obj}}^{(i)}$ is defined as $w_{\text{obj}}^{(i)} = (\sum_{j=1}^{N} \mathcal{K}_{\text{sim}}(i,j))/N,$
where {$\mathcal{{K}}_{\text{sim}}\in\mathbb{R}^{N\times N}$} represents the similarity matrix defined as $\mathcal{K}_{\text{sim}} = \mathbf{K} \mathbf{K}^\top$.

While Eq.~\eqref{eq:l_obj} aggregates the object-level features based on the \eigenmap{} assignment, we note that another kind of robust consistency could be cleverly imposed with our photometric augmented image $\Tilde{\mathbf{x}}$. That is, since the photometric augmentation does not apply structural changes, the augmented image $\Tilde{\mathbf{x}}$ and $\mathbf{x}$ are structurally identical, allowing us to make the following important assumption: the vectors in the same positions of $\mathbf{Z}$ and $\Tilde{\mathbf{Z}}$ should have similar object-level semantics.
This assumption ultimately allows us to create a new masked $\Tilde{\mathbf{Z}}$ (Fig.~\ref{fig:overview}, $\Tilde{\mathbf{Z}}$ in green box) of $\Tilde{\mathbf{x}}$ based on $\mathcal{M}_\text{eicue}$ of $\mathbf{x}$.
Thus, we apply the contrastive loss to the augmented image $\Tilde{\mathbf{x}}$, based on the prototypes $\Phi$ from the non-augmented image $\mathbf{x}$ to guide the model to learn global semantic consistency.
To illustrate this concept, our semantic consistency contrastive loss is defined as
\begin{equation}\small
\small \mathcal{L}_{\text{sc}}^{\mathbf{x}\rightarrow {\Tilde{\mathbf{x}}}} = \frac{1}{N} \sum_{i=1}^{N} w_{\text{obj}}^{(i)} \left[ -\log \left( \frac{\exp((\mathbf{\Tilde{Z}}_l^{(i)} \cdot \Phi_{l})/ \tau )}{\sum_{\substack{j=1, j\ne l}}^C \exp((\mathbf{\Tilde{Z}}_j \cdot \Phi_l )/ \tau)} \right) \right],
\end{equation}
where $\Tilde{\mathbf{Z}}_l^{(i)}$ notes the $i$-th feature vector of projected feature $\Tilde{\mathbf{Z}}$ for object $l$.
Concretely, we can formulate our object-centric contrastive loss as $\mathcal{L}_{\text{nce}}^{\mathbf{x}\rightarrow\Tilde{\mathbf{x}}} = \lambda_\text{obj}\mathcal{L}_{\text{obj}}^{\mathbf{x}\rightarrow {\mathbf{x}}} + \lambda_\text{sc}\mathcal{L}_{\text{sc}}^{\mathbf{x}\rightarrow\Tilde{\mathbf{x}}}$,
where $0<\lambda_\text{obj}<1$ and $0<\lambda_\text{sc}<1$ are hyperparameters that adjust the strength of each loss.
Since the loss function $\mathcal{L}_{\text{nce}}^{\mathbf{x}\rightarrow\Tilde{\mathbf{x}}}$ is asymmetric, we also take into account the opposite case as $\mathcal{L}_{\text{nce}}^{\Tilde{\mathbf{x}}\rightarrow \mathbf{x}} = \lambda_\text{obj}\mathcal{L}_{\text{obj}}^{\Tilde{\mathbf{x}}\rightarrow \Tilde{\mathbf{x}}} + \lambda_\text{sc}\mathcal{L}_{\text{sc}}^{\Tilde{\mathbf{x}}\rightarrow{\mathbf{x}}}$.
Therefore, the final \textit{object-centric contrastive loss} function~(ObjNCELoss) that we optimize is as follows:
\begin{equation}
\small \mathcal{L}_{\text{nce}}^{\mathbf{x}\leftrightarrow\Tilde{\mathbf{x}}} = \mathcal{L}_{\text{nce}}^{\mathbf{x}\rightarrow\Tilde{\mathbf{x}}} + \mathcal{L}_{\text{nce}}^{\Tilde{\mathbf{x}}\rightarrow \mathbf{x}}.
\end{equation}

\vspace{-7pt}
\subsection{Total Objective}
\vspace{-7pt}
To enhance the stability of the training process from the outset, we additionally employ a correspondence distillation loss~\cite{stego}, $\mathcal{L}_\text{corr}$
(see Supp D.1. for a detailed explanation).
In total, we minimize the following objective $\mathcal{L}_{\text{total}}$:
\begin{equation}
    \begin{aligned}
        \small \mathcal{L}_\text{total} = \lambda_\text{nce}\mathcal{L}_{\text{nce}}^{\mathbf{x}\leftrightarrow\Tilde{\mathbf{x}}} + (1-\lambda_\text{nce})\mathcal{L}_\text{corr} + \lambda_\text{eig}\mathcal{L}_{\text{eig}},
    \end{aligned}
    \label{eq:final}
\end{equation}
where $0\leqq\lambda_\text{nce}\leqq 1$ and $0\leqq\lambda_\text{eig}\leqq 1$ are hyperparameters. 
Here, $\lambda_\text{nce}$ starts from zero and increases rapidly, indicating the growing influence of $\mathcal{L}_{\text{nce}}^{\mathbf{x}\leftrightarrow\Tilde{\mathbf{x}}}$ during training.

\vspace{-5pt}
\section{Experiments}
\vspace{-5pt}

In this section, we first discuss the implementation details, including dataset configuration, evaluation protocols, and detailed experimental settings. 
Then, we evaluate our proposed method, \textit{\eagle{}}, both qualitatively and quantitatively while making a fair comparison with existing state-of-the-art methods. 
We also demonstrate the effectiveness of our proposed method through an ablation study.
See the supplementary material for additional details.

\vspace{-5pt}
\subsection{Experimental Settings}
\vspace{-5pt}
\myparagraph{Implementation Details.}
We use DINO~\cite{dino} pretrained vision transformer $\mathcal{F}$ which is kept frozen during the training process as in the prior works~\cite{stego, HP}. 
The training sets are resized and five-cropped to $244 \times 244$. 
For segmentation head $\mathcal{S}_\theta$, we use two layers of MLP with ReLU~\cite{stego, HP}, and for projection head $\mathcal{Z}_\xi$ we constructed a single linear layer~\cite{HP}.
All backbones employed an embedding dimension $D_{S}$ and $D_{Z}$ of 512.
For the \eigenmap{}, we extract 4 eigenvectors from the eigenbasis $\mathbf{V}$.
In the inference stage, we post-process the segmentation map with DenseCRF~\cite{crf, stego, HP}. 
See supplement for more details.

\begin{figure}[t!]
  \centering
  \includegraphics[width = \linewidth]{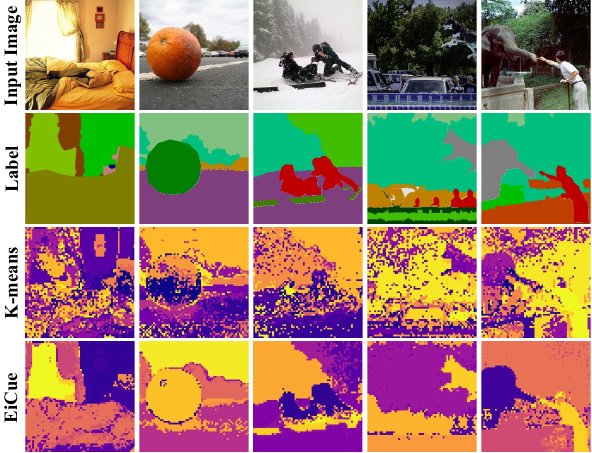}
\vspace{-20pt}
  \caption{
  Comparison between K-means and \eigenmap{}. 
  The bottom row presents \eigenmap{}, highlighting its superior ability to capture subtle structural intricacies and understand deeper semantic relationships, which is not as effectively achieved by K-means.
  }
\vspace{-17pt}
  \label{fig:eigenmaps}
\end{figure}

\myparagraph{Datasets.}
We evaluate on (1) COCO-Stuff~\cite{coco}, (2) Cityscapes~\cite{cityscapes}, and (3) Potsdam-3~\cite{iic} datasets, in line with methodologies established in prior works~\cite{HP, stego, picie, iic}. 
(1) The COCO-Stuff dataset is composed of its detailed pixel-level annotations, facilitating comprehensive various object understanding, while (2) Cityscapes presents diverse urban street scenes. (3) The Potsdam-3 dataset is composed of satellite imagery. Following the class selection protocols from previous studies ~\cite{HP, stego, picie, iic}, we use 27 classes from both COCO-Stuff and Cityscapes. For Potsdam-3, we use all 3 classes (see supplement for result).

\myparagraph{Evaluation Details.}
To align with established benchmarks, we adopt the evaluation protocols of prior works~\cite{stego, HP}. 
Our evaluation includes (1) a linear probe, assessing representational quality with a supervised linear layer on the unsupervised model, and (2) clustering through semantic segmentation via minibatch K-means based on cosine distance~\cite{clusteringprocess}, without ground truth, compared against it using Hungarian matching.
We measure performance using pixel accuracy (Acc.) and mean Intersection over Union (mIoU).

\begingroup
\setlength{\tabcolsep}{4.2pt} 
\renewcommand{\arraystretch}{1.0} 
\begin{table}[t!]
\caption{Quantitative results on the COCO-Stuff dataset~\cite{coco}.}
\vspace{-5pt}
    \centering
    \small
    \begin{tabular}{l c c c  c c}
        \hlineB{2.5}
        \multirow{2}{*}{Method}& \multirow{2}{*}{Backbone} & \multicolumn{2}{c}{Unsupervised} & \multicolumn{2}{c}{Linear} \\
        \multicolumn{2}{l}{}  & Acc. & mIoU & Acc. & mIoU \\
        \hline\hline
        DC~\cite{deepcluster} & R18+FPN & 19.9 & - & - & - \\
        MDC~\cite{deepcluster} & R18+FPN & 32.2 & 9.8 & 48.6 & 13.3 \\
        IIC~\cite{iic} & R18+FPN & 21.8 & 6.7 & 44.5 & 8.4 \\
        PiCIE~\cite{picie} & R18+FPN & 48.1 & 13.8 & 54.2 & 13.9 \\
        PiCIE+H~\cite{picie} & R18+FPN & 50.0 & 14.4 & 54.8 & 14.8 \\
        \hline
        SlotCon~\cite{wen2022slotcon} & R50 & 42.4 & 18.3 & - & - \\
        \hline
        DINO~\cite{dino} & ViT-S/16 & 22.0 & 8.0 & 50.3 & 18.1 \\ 
        + STEGO~\cite{stego} & ViT-S/16 & 52.5 & 23.7 & 70.6 & 34.5 \\ 
        + HP~\cite{HP} & ViT-S/16 & 54.5 & 24.3 & 74.1 & 39.1 \\
        \textbf{\textit{+ \eagle{}~(Ours)}} & ViT-S/16 & \textbf{60.1} & \textbf{24.4} & \textbf{75.2} & \textbf{42.5} \\
        
        \hline
        DINO~\cite{dino} & ViT-S/8 & 28.7 & 11.3 & 68.6 & 33.9 \\ 
        + TransFGU~\cite{transfgu} & ViT-S/8 & 52.7 & 17.5 & - & - \\
        + STEGO~\cite{stego} & ViT-S/8 & 48.3 & 24.5 & 74.4 & 38.3 \\
        + HP~\cite{HP} & ViT-S/8 & 57.2 & 24.6 & 75.6 & 42.7 \\ 
        \textbf{\textit{+ \eagle{}~(Ours)}} & ViT-S/8 & \textbf{64.2} & \textbf{27.2} & \textbf{76.8} & \textbf{43.9} \\ 
        
        \hlineB{2.5}
    \end{tabular}
    \label{Tab.coco}
\end{table}
\endgroup

\vspace{-3pt}
\subsection{Evaluation Results}
\vspace{-5pt}
Here, we carefully compare our proposed method to existing USS works in both qualitative and quantitative ways.
We mainly set up two representative baselines~\cite{stego, HP} from the literature which share the same evaluation protocols.

\begingroup
\setlength{\tabcolsep}{4.2pt} 
\renewcommand{\arraystretch}{1.0} 
\begin{table}[!t]
\caption{Quantitative results on the Cityscapes dataset~\cite{cityscapes}.}
\vspace{-5pt}
    \centering
    \small
    \begin{tabular}{l c c c  c c}
        \hlineB{2.5}
        \multirow{2}{*}{Method} & \multirow{2}{*}{Backbone} &
        \multicolumn{2}{c}{Unsupervised} & \multicolumn{2}{c}{Linear} \\
        \multicolumn{2}{l}{} & Acc. & mIoU & Acc. & mIoU \\
        \hline\hline
        MDC~\cite{deepcluster} & R18+FPN & 40.7 & 7.1 & - & - \\
        IIC~\cite{iic} & R18+FPN & 47.9 & 6.4 & - & - \\
        PiCIE~\cite{picie} & R18+FPN & 65.5 & 12.3 & - & - \\
        \hline
        DINO~\cite{dino} & ViT-S/8 & 34.5 & 10.9 & 84.6 & 22.8 \\
        + TransFGU~\cite{transfgu} & ViT-S/8 & 77.9 & 16.8 & - & - \\
        + HP~\cite{HP} & ViT-S/8 & 80.1 & 18.4 & \textbf{91.2} & 30.6 \\ 
        \textbf{+ \textit{\eagle{}~(Ours)}} & ViT-S/8 & \textbf{81.8} & \textbf{19.7} & \textbf{91.2} & \textbf{33.1}\\ 
        \hline
        DINO~\cite{dino} & ViT-B/8 & 43.6 & 11.8 & 84.2 & 23.0 \\
        + STEGO~\cite{stego} & ViT-B/8 & 73.2 & 21.0 & 90.3 & 26.8 \\
        + HP~\cite{HP} & ViT-B/8 & \textbf{79.5} & 18.4 & 90.9 & 33.0 \\ 
        \textbf{+ \textit{\eagle{}~(Ours)}} & ViT-B/8 & 79.4 & \textbf{22.1} & \textbf{91.4} & \textbf{33.4} \\ 
        \hlineB{2.5}
    \end{tabular}
\vspace{-10pt}
    
    \label{Tab.cityscapes}
\end{table}
\endgroup

\myparagraph{Quantitative Evaluation: COCO-Stuff.}
In Table~\ref{Tab.coco}, our \textit{\eagle{}} method sets new benchmarks on the COCO-Stuff dataset. 
\textbf{(I)} With the ViT-S/8 backbone, \textit{\eagle{}} showcases substantial improvements over existing methods in unsupervised accuracy, with gains of \textbf{+15.9} over STEGO~\cite{stego} and \textbf{+7.0} over HP~\cite{HP}. 
The unsupervised mIoU of \textit{\eagle{}} also significantly outperforms other methods: \textbf{+2.7} over STEGO and \textbf{+2.6} over HP.
The linear accuracy and mIoU of \textit{\eagle{}} both bring notable improvements over STEGO (\textbf{+2.4} Acc. and \textbf{+5.6} mIoU) and HP (\textbf{+1.2} Acc. and \textbf{+1.2} mIoU).
Compared to SlotCon~\cite{wen2022slotcon}, which also emphasizes object-level representations, our model excels with a \textbf{+21.8} and \textbf{+8.9} in unsupervised mIoU and accuracy respectively. 
\textbf{(II)} With the ViT-S/16 backbone, \textit{\eagle{}} maintains its dominance, gaining \textbf{+7.6} over STEGO and \textbf{+5.6} over HP in unsupervised Acc. The linear accuracy and mIoU of \textit{\eagle{}} outperforms STEGO (\textbf{+4.6} Acc. and \textbf{+8.0} mIoU) and HP (\textbf{+1.1} Acc. and \textbf{+3.4} mIoU) as well.

\myparagraph{Quantitative Evaluation: Cityscapes.}
As shown in Table~\ref{Tab.cityscapes}, our evaluations on the Cityscapes dataset show that \textit{\eagle{}} notably excels in both ViT-S/8 and ViT-B/8 backbones. 
\textbf{(I)} For the ViT-S/8 backbone, \textit{\eagle{}} 
has achieved significant unsupervised performance over STEGO (\textbf{+3.9} Acc. and \textbf{+2.9} mIoU) and HP (\textbf{+1.7} Acc. and \textbf{+1.3} mIoU). 
\textbf{(II)} For the ViT-B/8 backbone, \textit{\eagle{}} significantly improves both unsupervised Acc. and mIoU. 
The Cityscapes dataset innately exhibits highly imbalanced pixel-level class distributions, like the predominance of \texttt{sky} over \texttt{traffic light} pixels, typically forces a trade-off between Acc. and mIoU~\cite{HP}, as seen with STEGO and HP excelling in each metric respectively. However, \textit{\eagle{}} effectively balances these competing metrics, showcasing strong performance in both areas despite such challenges.

\myparagraph{Qualitative Analysis.}
In Fig.~\ref{fig:coco}, we also qualitatively compare our method to previous state-of-the-art models~\cite{stego, HP} on the COCO-Stuff and Cityscapes datasets trained using ViT-S/8 and ViT-B/8 backbone, respectively. 
Our approach outperforms baselines by accurately segmenting objects and preserving details, unlike STEGO which tends to segment multiple elements within a single object \texttt{furniture} or \texttt{road}, and HP neglects certain small objects \texttt{sports(kite)} or \texttt{traffic sign}. 
Our model, however, is trained at the object level with an understanding of the structure of the image, which not only comprehends the overall layout but also ensures no objects are missed.

\begingroup \small
\setlength{\tabcolsep}{2.9pt} 
\renewcommand{\arraystretch}{0.9} 
\begin{table}[t!]
\caption{Ablation results on the COCO-Stuff dataset~\cite{coco}.}
\vspace{-5pt}
    \centering
    \small 
    \begin{tabular}{c|c c cc c cc|cc|cc}
    \hlineB{2.5}
        {\footnotesize Exp.} & \multicolumn{1}{c}{\multirow{2}{*}{$\mathcal{L}_{\text{corr}}$}} & & \multicolumn{2}{c}{$\mathbf{x} \rightarrow \Tilde{\mathbf{x}}$} & & \multicolumn{2}{c|}{$\Tilde{\mathbf{x}} \rightarrow \mathbf{x}$} & \multicolumn{1}{c}{\multirow{2}{*}{$\mathcal{M}_{\text{eicue}}$}} & \multicolumn{1}{c|}{\multirow{2}{*}{$\mathcal{M}_{\text{km}}$}} & \multicolumn{2}{c}{\footnotesize Unsupervised} \\
        
        \cline{4-5}\cline{7-8}
        \# & & & $\mathcal{L}_{\text{obj}}$ & $\mathcal{L}_{\text{sc}}$ & & $\mathcal{L}_{\text{obj}}$ & $\mathcal{L}_{\text{sc}}$ & & & {\footnotesize Acc.} & {\footnotesize mIoU} \\
    \hlineB{2.5}
        
        1 & \checkmark   & &  &  & &  & & & & 46.9 & 21.8 \\      
        2 & \checkmark   & &  & \checkmark & &  & \checkmark & \checkmark & & 59.3 & 23.2  \\
        3 & \checkmark   & & \checkmark & & & \checkmark & & \checkmark & & 62.1 & 25.1  \\
        4 & \checkmark   & &  &  & & \checkmark & \checkmark & \checkmark & & 61.6 & 24.8 \\
        5 & \checkmark   & & \checkmark & \checkmark & & & & \checkmark & & 62.9 & 26.1 \\
        6 & \checkmark   & & \checkmark & \checkmark & & \checkmark & \checkmark & & \checkmark & 55.1 & 17.0 \\
    \hlineB{1.5}
        7 & \checkmark   & & \checkmark & \checkmark  & & \checkmark & \checkmark & \checkmark & & \textbf{64.2} & \textbf{27.2} \\
    \hlineB{2.5}
    \end{tabular}
\vspace{-13pt}
    \label{tab:ablation}
\end{table}
\endgroup

\vspace{-5pt}
\subsection{Ablation Study} \label{sec:ablation}
\vspace{-7pt}
We further analyze our model with ablation studies and discuss the results based on the full ablation results in Table~\ref{tab:ablation} denoted as Exp.~\#1 to Exp.~\#7.
We primarily conducted our experiments using the COCO-Stuff dataset using the DINO pretrained ViT-S/8 model. 
For more details, please refer to the supplementary material.

\begin{figure}[t!]
  \centering
  \includegraphics[width = \linewidth]{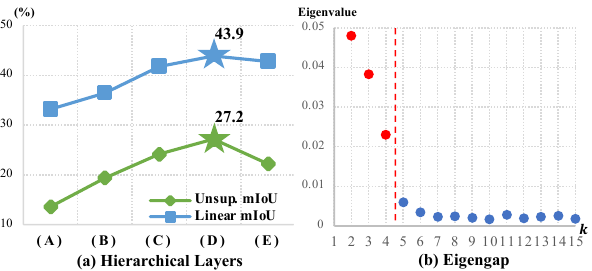}
\vspace{-17pt}
  \caption{\textbf{(a)} Analysis of hierarchical attention with the following layer combinations (layer numbers in square brackets): (A): [1-6-12], (B): [12], (C): [11-12], (D): [10-11-12], and (E): [9-10-11-12]. \textbf{(b)} Analysis of eigengap to identify the optimal $k$ for eigenbasis clustering, selected at the dashed line with maximal eigengap (i.e., the gap between two consecutive eigenvalues).
  }
\vspace{-18pt}
  \label{fig:hierarchy}
\end{figure}

\myparagraph{Effect of \eigenmap{}.}
We validate the effectiveness of \eigenmap{}~($\mathcal{M}_{\text{eicue}}$) by comparing the performance of our \eigenmap{}-enhanced method (Exp.~\#7) against a K-means~($\mathcal{M}_{\text{km}}$) approach (Exp.~\#6) in Table~\ref{tab:ablation}. The \eigenmap{} result shows a notable improvement, capturing fine structural details that K-means misses. Fig.~\ref{fig:eigenmaps} visually demonstrates how \textit{\eagle{}} better identifies object semantics and structures compared to K-means.

\myparagraph{ObjNCE Loss.}
Table~\ref{tab:ablation} shows how different loss components affect performance. The full model (Exp.~\#7) outperforms others, highlighting the effectiveness of combining all components. Notably, using $\mathcal{L}_{\text{obj}}$ alone (Exp.~\#3) significantly improves upon the baseline, underlining the importance of object-focused representation. The inclusion of $\mathcal{L}_{\text{sc}}$ further refines quality, as evidenced by comparing Exp.~\#3 with Exp.~\#7. Additionally, the combined use of both $\mathcal{L}_{\text{nce}}$ directions (Exp.~\#7) shows a synergistic effect over using them individually (Exp.~\#4 and Exp.~\#5).

\myparagraph{Combination of Hierarchical Attention and Eigengap.}
In Fig.~\ref{fig:hierarchy}a, we present results from using various combinations of hierarchical attention. 
The combination of the third-to-last, second-to-last, and last layers from 12-layer architecture, demonstrated the best performance since the layers closer to the end better capture the spatial information of the image.
For optimal eigenbasis clustering, we conduct eigengap analysis in Fig.~\ref{fig:hierarchy}b. Since we choose $k$ at the point where the eigengap is maximized, we have selected $k=4$.

\let\thefootnote\relax\footnote{\scriptsize{{\bf Acknowledgement.} This work was supported in part by the National Research Foundation of Korea (NRF) Grant funded by the Korean Government through the Ministry of Science and ICT (MSIT) under Grant RS-2023-00219019, and Institute of Information \& Communications Technology Planning \& Evaluation (IITP) Grant funded by MSIT (No. 2020-0-01361, Artificial Intelligence Graduate School Program (Yonsei University)).}}

\vspace{-18pt}
\section{Conclusion}
\vspace{-7pt}
In this study, we present \textit{EAGLE}, a novel method that addresses the persistent challenges in semantic segmentation with a focus on collecting semantic pairs through an object-centric lens.
Through empirical analysis using a series of datasets, \textit{EAGLE} showcases a remarkable capability to leverage the Laplacian matrix constructed from attention-projected features and fortified by an object-level prototype contrastive loss, which guarantees the accurate association of objects with their corresponding semantic pairs. 
Pioneering in utilizing dual advanced techniques, this method marks a substantial advance in addressing the constraints of patch-level representation learning found in previous research. 
Consequently, \textit{EAGLE} emerges as a powerful framework for encapsulating the semantic and structural intricacies of images in contexts devoid of labels.

{
    \small
    \bibliographystyle{ieeenat_fullname}
    \bibliography{main}
}

\clearpage

\setcounter{page}{1}

\section*{A. Additional Material: Project Page \& Presentation Video}
We have described our results in an easily accessible manner on our project page, where a brief \textbf{presentation video} is also available. The link to the \textbf{project page} is as follows: \url{https://micv-yonsei.github.io/eagle2024/}.

\section*{B. Additional Evaluation Results}
In this section, we extend our discussion to include the evaluation results of \textit{EAGLE}. Initially, we present both quantitative and qualitative findings from the \textbf{Potsdam-3} dataset~\cite{iic}, which were not covered in the main paper due to space constraints. Subsequently, we also provide additional qualitative analysis of the COCO-Stuff~\cite{coco} and Cityscapes datasets~\cite{cityscapes}.

\subsection*{B.1. Potsdam-3}

\begingroup
\setlength{\tabcolsep}{10pt} 
\renewcommand{\arraystretch}{1.0} 
\begin{table}[h!]
\caption{Quantitative results on Potsdam-3 dataset~\cite{iic}.}
    \centering\resizebox{\linewidth}{!}{
    \small
    \begin{tabular}{l c c c}
        \hlineB{2.5}
        Method & Backbone & Unsup. Acc. & Unsup. mIoU \\
        \hline\hline
        Random CNN~\cite{iic}& VGG11 & 38.2 & - \\
        K-means~\cite{sklearn}& VGG11 & 45.7 & -  \\
        SIFT~\cite{sift} & VGG11 &38.2 & -  \\
        ContextPrediction~\cite{context} & VGG11 & 49.6 & -  \\
        CC~\cite{isola2015learning} & VGG11 & 63.9 & -  \\
        DeepCluster~\cite{deepcluster} & VGG11 & 41.7 & -  \\
        IIC~\cite{iic} & VGG11 & 65.1 & -  \\
        \hline
        DINO~\cite{dino} & ViT-B/8 & 53.0 & -  \\
        + STEGO~\cite{stego}& ViT-B/8 & 77.0 & 62.6  \\
        + HP~\cite{HP} & ViT-B/8 & 82.4 & 68.6 \\
        \textbf{+ \textit{EAGLE~(Ours)}} & ViT-B/8 & \textbf{83.3} & \textbf{71.1}\\
        \hlineB{2.5}
    \end{tabular}}
    
    \label{Tab.potsdam}
\end{table}
\endgroup

In Table~\ref{Tab.potsdam}, we present the quantitative results for the Potsdam-3 dataset~\cite{iic}, where our \textit{EAGLE} sets a new score.
We not only report the unsupervised accuracy, as previously done by methods~\cite{stego, HP}, but also expand our reporting to include unsupervised mIoU.
With the ViT-B/8 backbone, \textit{EAGLE} surpass existing USS methods in unsupervised accuracy, with gains of \textbf{+6.3} over STEGO~\cite{stego} and \textbf{+0.9} over HP~\cite{HP}.
In the context of unsupervised mIoU, our method surpasses STEGO by a significant margin of \textbf{+8.5}, and \textbf{+2.5} over HP.

\begin{figure}[h!]
  \centering
  \includegraphics[width = \linewidth]{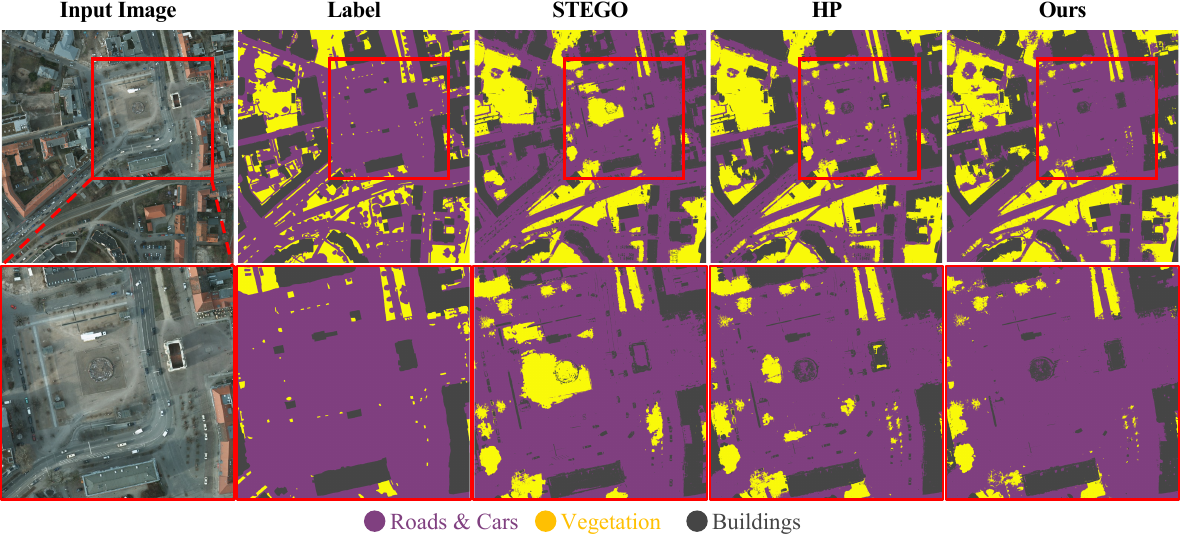}
  \caption{Qualitative results of Potsdam-3 dataset~\cite{iic} trained with ViT-B/8 backbone.
  }
  \label{fig:potsdam_supp}
\end{figure}

In our qualitative analysis in Fig.~\ref{fig:potsdam_supp}, compared to STEGO~\cite{stego} and HP~\cite{HP}, our \textit{EAGLE} demonstrates a more accurate understanding of object-level semantics. 
Specifically, in the second row, which is a zoomed-in view of the red box in the first row, \textit{EAGLE} successfully classifies the cars on the road as separate entities from the buildings. 
This distinction is not as clear in the results from STEGO and HP, highlighting the superior capability of our approach in discerning and segmenting objects according to their semantic categories.

\subsection*{B.2. COCO-Stuff}
We demonstrate additional qualitative results in Fig.~\ref{fig:coco_supp}.

\subsection*{B.3. Cityscapes}
Additional qualitative results are available in Fig.~\ref{fig:city_supp}.

\section*{C. Additional Experiments}
\subsection*{C.1. Additional Ablation Study}
In this section, we provide additional ablation analysis on the feature type~(Section C.1.1) and prototype selection method~(Section C.1.2).

\subsubsection*{C.1.1 Ablation: Feature Type}
\label{sec:featuretype}

\begingroup
\setlength{\tabcolsep}{15pt} 
\renewcommand{\arraystretch}{1.0}
\begin{table}[!h]
\caption{The experimental results for the feature type on the COCO-Stuff dataset.}
    \centering\resizebox{\linewidth}{!}{
    \small
    \begin{tabular}{c c|c c|c c}
        \hlineB{2.5}
        \multicolumn{2}{c|}{Feature Type} & \multicolumn{2}{c|}{Unsupervised} & \multicolumn{2}{c}{Linear} \\
        \multicolumn{1}{c}{\multirow{1}{*}{$\mathcal{S}_\theta$}}          & $\mathbf{A}_\text{seg}$          & Acc.        & mIoU         & Acc.        & mIoU        \\ \hline\hline
        \multicolumn{1}{c|}{\multirow{3}{*}{$\mathbf{F}$}} & $\mathbf{F}$ & 43.1       & 17.1         & 74.1       & 41.2        \\
        \multicolumn{1}{c|}{}                    & $\mathbf{K}$ & 57.6        & 24.9          &  74.6          & 41.6            \\
        \multicolumn{1}{c|}{}                    & $\mathcal{S}_\theta(\mathbf{F})$ & 59.1         & 25.4          & 74.5           &  41.5           \\ \hline
        \multicolumn{1}{c|}{\multirow{3}{*}{$\mathbf{K}$}} & $\mathbf{F}$ & 58.6       &  26.1        & 74.7       & 41.7             \\
        \multicolumn{1}{c|}{}                    & $\mathbf{K}$ & 56.9       &  23.8        & 74.6       & 41.6         \\
        \multicolumn{1}{c|}{}                    &  $\mathcal{S}_\theta(\mathbf{K})$ & \textbf{64.2} & \textbf{27.2} & \textbf{76.8} & \textbf{43.9} \\ \hlineB{2.5}
    \end{tabular}}
    \label{Tab.featuretype}
    \end{table}
\endgroup

In this section, we explore various different combinations of feature types that will be used for computing $\mathcal{S}_\theta$ and for the creation of $\mathbf{A}_\text{seg}$, thereby exploring their implications and potential applications in the context of our study.
In Table~\ref{Tab.featuretype}, we analyze the experimental results for the COCO-Stuff dataset~\cite{coco} trained using the ViT-S/8 backbone. 
The $\mathbf{F}$ and $\mathbf{K}$ listed under the $\mathcal{S}_\theta$ in the ``Feature Type" column, represent the types of features inputted into the segmentation head $\mathcal{S}_\theta$. 
In this study, $\mathbf{F}$ is sourced from the activation map at the final layer of the vision transformer, while $\mathbf{K}$ is processed in accordance with the methods outlined in the main manuscript (Section 3.1).
The $\mathbf{A}_\text{seg}$ located in the ``Feature Type" column, denotes the feature types utilized in the creation of $\mathbf{A}_\text{seg}$. 
As detailed in the manuscript, EiCue construction involves the sum of two adjacency matrices to form the Laplacian, one of which is the semantic similarity matrix $\mathbf{A}_\text{seg}$, providing the semantic interpretation of the object. 
The $\mathbf{F}$ and $\mathbf{K}$ in the $\mathbf{A}_\text{seg}$ column, retain the same values as previously described, being sourced from a static pretrained vision transformer. 
When considering $\mathcal{S}_{\theta}(\mathbf{F})$ and $\mathcal{S}_{\theta}(\mathbf{K})$, these refer to the $\mathbf{F}$ and $\mathbf{K}$ features that have been processed through the segmentation head $\mathcal{S}_\theta$. 
As the model training progresses, these dynamic values are subject to change, reflecting the evolving state of the trained segmentation head.

As detailed in Table~\ref{Tab.featuretype}, we present the results for all possible feature type combinations. 
Overall, leveraging $\mathbf{K}$ to compute $\mathbf{S}$ through $\mathcal{S}_\theta$, yielded superior outcomes compared to utilizing $\mathbf{F}$. 
To construct $\mathbf{A}_\text{seg}$, utilizing dynamic feature types such as $\mathcal{S}_{\theta}(\mathbf{F})$ or $\mathcal{S}_{\theta}(\mathbf{K})$ demonstrates significantly higher values in contrast to the aforementioned static features, $\mathbf{F}$ and $\mathbf{K}$, thereby validating the effectiveness of our training approach. 
Ultimately, employing the learnable feature $\mathcal{S}_\theta(\mathbf{K})$ for $\mathbf{A}_\text{seg}$ delivered the best performance.

\subsubsection*{C.1.2 Ablation: Prototype Selection Method}
\label{sec:prototypes}

\begingroup
\setlength{\tabcolsep}{15pt} 
\renewcommand{\arraystretch}{1.0}
\begin{table}[!h]
\caption{The experimental results for the prototype selection method on the COCO-Stuff dataset.}
    \centering\resizebox{\linewidth}{!}{
    \small
    \begin{tabular}{c|cc|cc}
    \hlineB{2.5}
    \multirow{2}{*}{Method} & \multicolumn{2}{c|}{Unsupervised} & \multicolumn{2}{c}{Linear}\\
                            & Acc.            & mIoU           & Acc.         & mIoU        \\
    \hline\hline
    PCA         &            55.7     &         18.3       &       74.7       &      39.6       \\
    Centroid                    &  59.0               &      24.9          &       75.8       &      42.1       \\
    Medoid                  & \textbf{64.2} & \textbf{27.2} & \textbf{76.8} & \textbf{43.9}             \\
    \hlineB{2.5}
    \end{tabular}}
    \label{Tab.prototype}
    \end{table}
\endgroup

We further carry out ablation experiments comparing various methods for selecting prototypes.
A prototype is a semantic vector that represents a single object, serving as an anchor that attracts semantic vectors within the object and repels the other ones.
To select a semantic vector that represents an object, we can consider several options for choosing object-representative semantic vectors.
\textbf{(a)} Principal Component Analysis (PCA): we can use PCA to find the direction of maximum variance in the data, and choose the vector that has the largest projection on the first principal component.
\textbf{(b)} Centroid: calculates the mean vector of the set, where each component of the centroid is the average of that component across all vectors in the set.
\textbf{(c)} Medoid: we can choose the vector that minimizes the sum of distances to all other vectors. This is less sensitive to outliers compared to the centroid.
As reported through Table~\ref{Tab.prototype}, we observed that leveraging the Medoid method demonstrated the best performance.

\subsection*{C.2. Additional Visualization of the Primary Elements of Eigen Aggregation Module}
\subsubsection*{C.2.1 Eigenvectors}

As illustrated in Fig.~\ref{fig:eigne_supp}, we provide additional visualization of eigenvectors obtained from learnable feature $\mathbf{S}$ trained using COCO-Stuff dataset~\cite{coco}.
Our eigenvectors present remarkable capability in distinguishing objects while capturing within its object semantics. 

\subsubsection*{C.2.2 EiCue}

As shown in Fig.~\ref{fig:eicue_supp}, we present additional visualizations that compare our EiCue model with the traditional K-means clustering approach.
EiCue shows a significant improvement over clustering with K-means in identifying the semantic details of objects and discerning the comprehensive structure of images. This distinction is particularly evident in the way EiCue captures intricate object semantics and delineates the structural elements within the images. These observations substantiate our model's claim that EiCue is proficient in recognizing object semantics and distinguishing structural components.

\subsection*{C.3. Application}
Our \textit{EAGLE} is designed for semantic segmentation, which predicts dense class prediction. 
Consequently, \textit{EAGLE} is applicable to a variety of tasks that necessitate pixel-level semantic interpretation. 
A key technique in our approach is the \textit{eigendecomposition} of the Laplacian matrix. 
Thus, we can use eigenvectors of images and these eigenvectors are able to capture the detailed semantic structures present in an image.
As discussed in Section C.3.1, leveraging these eigenvectors allows us to precisely perform image matting and achieve localized image stylization.

\subsubsection*{C.3.1 Image Matting}
\label{sec:matting}

\begin{figure}[t]
  \centering
  \includegraphics[width = \linewidth]{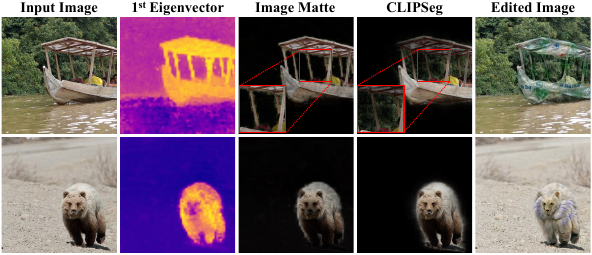}
  \begin{minipage}{\linewidth} 
    \caption{We demonstrate our Laplacian matrix-based image matting. From left to right: the input image, eigenvector from our matrix, resultant image matte, CLIP-based segmentation~\cite{lueddecke22_cvpr}, and edited image using our matte with existing text-driven image editing model~\cite{lee2022lisa}. For the editing in the last column, the text prompts used were \texttt{plastic bag} for the top image and \texttt{white fur} for the bottom image. Our matte offers clearer object boundaries than CLIPSeg, leading to superior editing quality.}
    \label{fig:matting}
  \end{minipage}
\end{figure}

The eigenvectors obtained through our proposed method are effective in distinguishing objects within images. 
To this end, our eigenvectors can be efficiently utilized for image matting and localized image stylization.
The process of image matting involves the extraction of the foreground from an image, facilitating further manipulations like compositing onto a different background or selective editing. Historically, the matting task has been challenging due to intricate object edges and subtle transitions. The eigenvectors, adept at distinguishing distinct objects or features within images, provide a powerful solution to this challenge.
Using our proposed method, we leverage the potential of these eigenvectors for refined image matting. 
As depicted in Fig.~\ref{fig:matting}, the first column represents the input images. The subsequent column showcases the first eigenvector derived from our matrix, effectively highlighting the structure of the primary object. The third column portrays the resultant image matte, distinctly separating the object from its surroundings. In contrast, the fourth column, representing the CLIP-based segmentation, while reasonable, but fails to provide as delicate boundaries as our eigenvector-based technique. A notable difference can be observed in the image of the \texttt{boat}. Our method adeptly separates the pillar of the boat, whereas the CLIP-based approach fails to isolate it, erroneously including the trees in the background as part of the foreground. 
The final column presents the edited images, emphasizing the utility of our matte for selective edits, ensuring that the distinguished object can be stylized without affecting the background.

\section*{D. Implementation Details}
In this section, we discuss details of correspondence distillation loss (Section D.1.), model architecture (Section D.2.), and hyperparameters (Section D.3).

\subsection*{D.1. Correspondence Distillation Loss}
\label{sec:cdl}

By employing a correspondence distillation loss~\cite{stego}, we enhanced the stability of the training process by ensuring reliable graph Laplacian initialization. 
The original correspondence distillation loss is defined as 
\begin{equation}
    \begin{aligned}
    \mathcal{L}_\text{cd}(\check{\mathbf{F}}, \check{\mathbf{S}}, b) = -\sum \left(\check{\mathbf{F}} -b\right)\check{\mathbf{S}},
   \end{aligned}
\label{eq:cd}
\end{equation}
where $\check{\mathbf{F}}$ and $\check{\mathbf{S}}$ is computed as a cosine distance using $\mathbf{F}$ and $\mathbf{S}$.
Here, $\mathbf{F}$ is a feature obtained from the activation map at the final layer of the vision transformer with a given image and $\mathbf{S}$ is the projection of $\mathbf{F}$ using segmentation head $\mathcal{S}_\theta$. 
Since we leverage attention keys in place of $\mathbf{F}$, we substitute $\mathbf{F}$ with $\mathbf{K}$ and revise $\mathbf{S}$ to be $\mathbf{S}=\mathcal{S}_\theta(\mathbf{K})$.
Within the framework of existing correspondence distillation loss~\cite{stego}, which involves three distinct loss functions, our method modifies and utilizes two of these components: \textbf{(a)} the augmented image correspondence distillation loss and the \textbf{(b)} random image correspondence distillation loss.

Although Eq.~\eqref{eq:cd} is applied to both types of loss, the difference lies in what each correspondence tensor represents. 
\textbf{(a)} In the augmented image correspondence distillation loss, $\mathbf{\check{K}}_{\text{aug}}$ and $\mathbf{\check{S}}_{\text{aug}}$ is computed as a cosine distance between $\mathbf{K}, \mathbf{\Tilde{K}}$ and $\mathbf{S}, \mathbf{\Tilde{S}}$, respectively. $\mathbf{\Tilde{K}}$ and $\mathbf{\Tilde{S}}$ are the results for the $\mathbf{\Tilde{x}}$, which is the augmented images of $\mathbf{x}$, created through the same aforementioned process. While \textbf{(b)} in the random image correspondence distillation loss, $\mathbf{\check{K}}_{\text{rand}}$ and $\mathbf{\check{S}}_{\text{rand}}$ is computed as a cosine distance between $\mathbf{K}, \mathbf{\acute{K}}$ and $\mathbf{S}, \mathbf{\acute{S}}$, respectively. $\mathbf{\acute{K}}$ and $\mathbf{\acute{S}}$ are the results for the random images from the entire dataset, created through the same aforementioned process.
In the Eq.~\eqref{eq:cd}, $b$ is defined as the \textit{shift} of the feature value and remained fixed throughout the training process.
In contrast, we modified $b_{\text{aug}}$ and $b_{\text{rand}}$ to dynamically adapt based on the $\check{\mathbf{K}}$ and $\check{\mathbf{S}}$ in both losses, where $b_{\text{aug}}$ represents the $b$ in augmented image correspondence distillation loss, and $ b_{\text{rand}}$ represents the $b$ in random image correspondence distillation loss. Here is the formula we used:

\begin{equation}
    \begin{aligned}
    \small
    b_{\text{aug}} = \left\lvert {\frac{1}{HW} \sum_{i=1}^{H}\sum_{j=1}^{W}\check{\mathbf{K}}_{ij}} - {\frac{1}{HW} \sum_{i=1}^{H}\sum_{j=1}^{W}\check{\mathbf{S}}_{ij} - k_{\text{shift}}} \right\rvert, 
   \end{aligned}
\label{eq:final}
\end{equation}

\begin{equation}
\small
    \begin{aligned}
    b_{\text{rand}} = \left ({\frac{1}{HW} \sum_{i=1}^{H}\sum_{j=1}^{W}\check{\mathbf{K}}_{ij}} + {\frac{1}{HW} \sum_{i=1}^{H}\sum_{j=1}^{W}\check{\mathbf{S}}_{ij}}- k_{\text{shift}}\right) \times v_{\text{shift}},
   \end{aligned}
    \label{eq:final}
\end{equation}
where $H$ and $W$ refer to the height and width of the feature tensor, respectively. Here, $k_\text{shift}$ and $v_\text{shift}$ are determined as hyperparameters (see Section D.3.)
Thus, the final correspondence distillation loss that we use is defined as
\begin{equation}
    \begin{aligned}
    \small
    \mathcal{L}_\text{corr} =  \mathcal{L}_\text{cd}(\check{\mathbf{K}}_\text{aug}, \check{\mathbf{S}}_\text{aug}, b_\text{aug}) + \mathcal{L}_\text{cd}(\check{\mathbf{K}}_\text{rand}, \check{\mathbf{S}}_\text{rand}, b_\text{rand}),
   \end{aligned}
\label{eq:final}
\end{equation}
which is the summation of augmented image correspondence distillation loss and random image correspondence distillation loss.

\subsection*{D.2. Detailed Architecture}
\label{sec:detailarchitecture}

\begin{figure}[h!]
  \centering
  \includegraphics[width = \linewidth]{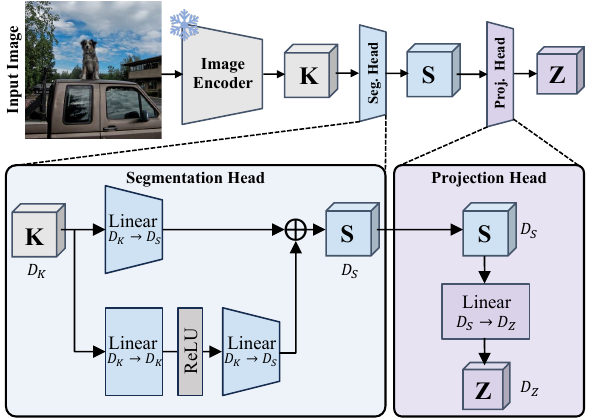}
  \caption{Detailed architecture of segmentation head and projection head used in our method.
  }
  \label{fig:seghead}
\end{figure}

\myparagraph{Image Encoder.} For all experiments, we basically leverage DINO~\cite{dino} pretrained ViT-S/8, ViT-S/16 and ViT-B/8~\cite{vit} as an image encoder. 
Specifically, we initialize ViT with a teacher weight of DINO.
As mentioned before, we extract attention keys hierarchically from ViT and then concatenate them into a single feature tensor $\mathbf{K}$ (for details, see Section 3.1 in the main manuscript). 
Then, we apply channel-wise dropout ($p = 0.1$) to feature tensor $\mathbf{K}$ before feeding to segmentation head $\mathcal{S}_\theta$. 

\myparagraph{Segmentation Head.} We illustrate the detailed architecture of the segmentation head and projection head in Fig.~\ref{fig:seghead}. 
For a fair performance comparison with existing models, we employ the same approach for the segmentation head $\mathcal{S}_\theta$ as used in the previous models~\cite{stego, HP}.
This non-linear segmentation head $\mathcal{S}_\theta$ consists of simple linear layers.
The input is a tensor $\mathbf{K}$ with a dimension of $D_K$.
This tensor first passes through a linear layer that transforms its dimension from $D_K$ to $D_S$, where $D_S$ represents the desired dimension for the output of $\mathcal{S}_\theta$.
Following the initial linear transformation, there is a ReLU (Rectified Linear Unit) activation function, which introduces non-linearity to the process. The output of the activation layer is then fed into another linear layer, which once again maps the dimension from $D_K$ to $D_S$.
The outputs of the two pathways are then combined via a summation operation. 
The summation consolidates the linearly transformed input and the non-linearly transformed input. The result is the tensor $\mathbf{S}$ with the dimension $D_S$, which is the output of the segmentation head.

\myparagraph{Projection Head.}
As shown in Fig.~\ref{fig:seghead}, we project semantic tensor $\mathbf{S}$ to $\mathbf{Z}$ to facilitate object-centric contrastive learning.
The basic concept of projection head $\mathcal{Z}_\xi$ is to project the tensor without transforming its input dimension.
Following this concept, we form a projection head with a single linear layer that maps its dimension from $D_S$ to $D_Z$.
Here, we note that while we use different notations $D_S$ and $D_Z$ for clarity and ease of explanation, the actual dimensions represented by these notations are the same as $D_S=D_Z$.

\subsection*{D.3. Hyperparameters}
\label{sec:hyper}

\begingroup
\begin{table}[h]
\caption{Hyperparameters used in \textit{EAGLE}. LR refers learning rate.}
\centering\resizebox{\linewidth}{!}{
\begin{tabular}{c|c|c|c}

\hlineB{2.5}
\multicolumn{2}{c|}{Hyperparams}  & COCO-Stuff ViT-S/8 & Cityscapes ViT-B/8  \\
\hline\hline
\multicolumn{2}{c|}{$\lambda_{\text{obj}}$} &   0.3   &   0.3    \\
\multicolumn{2}{c|}{$\lambda_{\text{sc}}$}  &   0.7   &    0.7    \\
\multicolumn{2}{c|}{$\lambda_{\text{nce}}$} &    0.9  &    0.7   \\
\multicolumn{2}{c|}{$k_\text{shift}$}         &   0  &     0.11   \\
\multicolumn{2}{c|}{$v_\text{shift}$}         &   3.5  &     3.5   \\
\multicolumn{2}{c|}{step}                   &   200  &     380   \\
\hline
\multirow{2}{*}{LR}   & $\mathcal{S}_\theta , \mathcal{Z}_\xi , \Phi$ &  0.0005    &  0.0005   \\
                      & $\mathbf{C}$     &   0.00005   &   0.0004  \\
\hlineB{2.5}
\end{tabular}}
\label{Tab.hyperparameter}
\end{table}
\endgroup

In this section, we carefully describe hyperparameters in Table~\ref{Tab.hyperparameter} that are used throughout our series of experiments.
In the table above, ``step" refers to the number of training iterations required for the $\lambda_{\text{nce}}$ to increase from 0 to the number indicated in $\lambda_{\text{nce}}$ row. 
``LR" indicates learning rate and $\mathcal{S}_\theta, \mathcal{Z}_\xi, \Phi$ shares same learning rate.

\section*{E. Discussion}
\subsection*{E.1. Failure Cases}

\begin{figure}[h!]
  \centering
  \includegraphics[width = \linewidth]{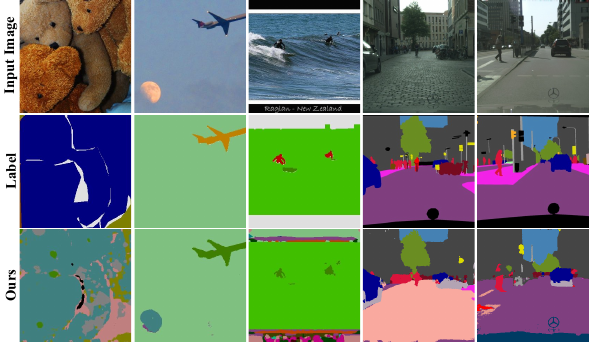}
  \begin{minipage}{.7\linewidth}
  \caption{Failure cases of \textit{EAGLE}.
  }
  \label{fig:failure}
  \end{minipage}
\end{figure}

Unsupervised semantic segmentation (USS), unlike the fully-supervised approach, is quite challenging as it predicts classes for each pixel without labeled data.
While USS is likely to show much better performance in the future, our model represents a step in its evolution and thus comes with certain limitations.
Fig.~\ref{fig:failure} illustrates the failure cases of EAGLE, which has been trained to capture object-level semantics.
The first to the third column is from the COCO-Stuff dataset and the remaining columns are from the Cityscapes dataset.
In the first column, our model fails to segment objects properly.
This is due to the narrow color distribution of the input image and the limited variety of object semantics present in the image, leading to a failure in creating a high-quality adjacency matrix for EiCue.
In the second and third columns, we observed that our results successfully implemented object-level semantics but made errors in matching the object class.
Within column four, we see our results that accurately segment and correctly classify \texttt{car}, \texttt{tree}, \texttt{building}, and \texttt{person}, but incorrectly categorize gravel paths as a different class instead of \texttt{road}.
Similarly, in the last column, there were no critical errors for objects other than \texttt{sidewalk}. 
However, even when viewed with the human eye, the input image presents a challenging scenario in distinguishing between \texttt{road} and \texttt{sidewalk}.

\subsection*{E.2. Future Works}

Throughout our manuscript, we demonstrated that leveraging EiCue through graph Laplacian effectively captures the semantic structure of an image. 
However, constructing an adjacency matrix and forming a Laplacian matrix entails a relatively high computational cost.
This approach does not affect the inference time in our framework, but it does require more training time compared to using solely deep-based methods.
In our research, we compute the adjacency matrix for every feature vector of an image. 
While \textit{EAGLE} shows state-of-the-art results, regarding every single feature vector is not the most computationally efficient, suggesting that improvements to EiCue could be made by sampling only vital features based on other knowledge of the image and constructing the Laplacian matrix accordingly.
Additionally, as our primary focus is on object-level semantics, this approach may not be directly applicable to domains like medical imaging.
Therefore, it is crucial to engage in research that uncovers knowledge about object-level semantics, which is applicable across multiple domains and holds significant potential for widespread use in the field of computer vision.

\begin{figure*}[t!]
  \centering
  \includegraphics[width = 0.95\linewidth]{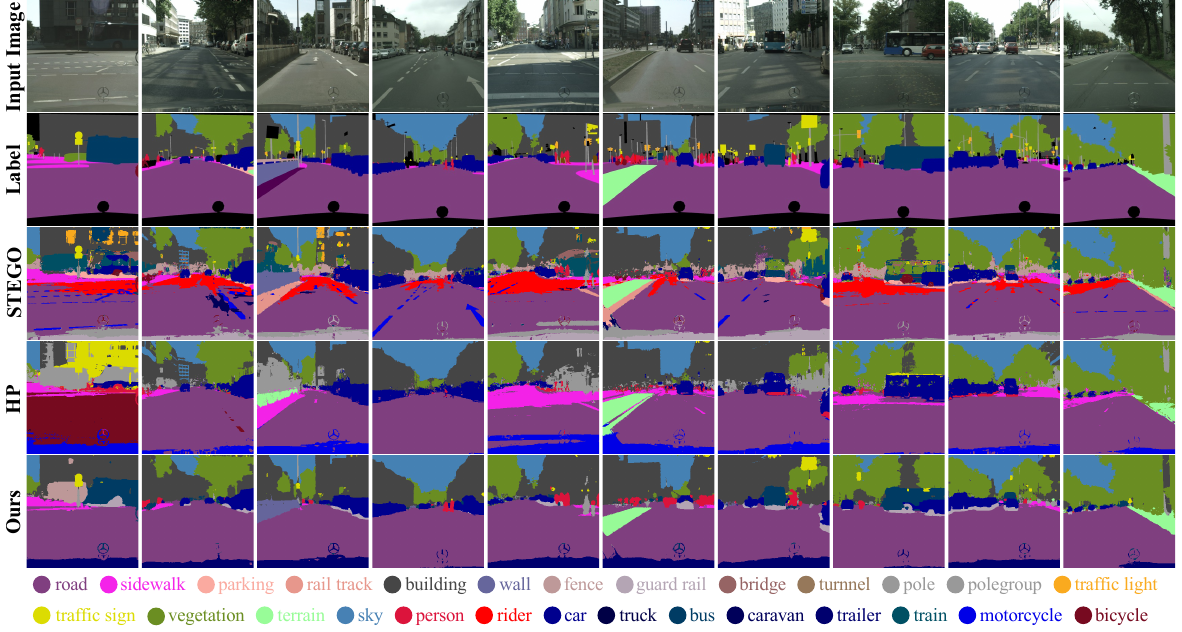}
  \caption{Additional qualitative results of Cityscapes dataset~\cite{cityscapes} trained with ViT-B/8 backbone.
  }
  \label{fig:city_supp}
\end{figure*}

\begin{figure*}[t!]
  \centering
  \includegraphics[width = 0.95\linewidth]{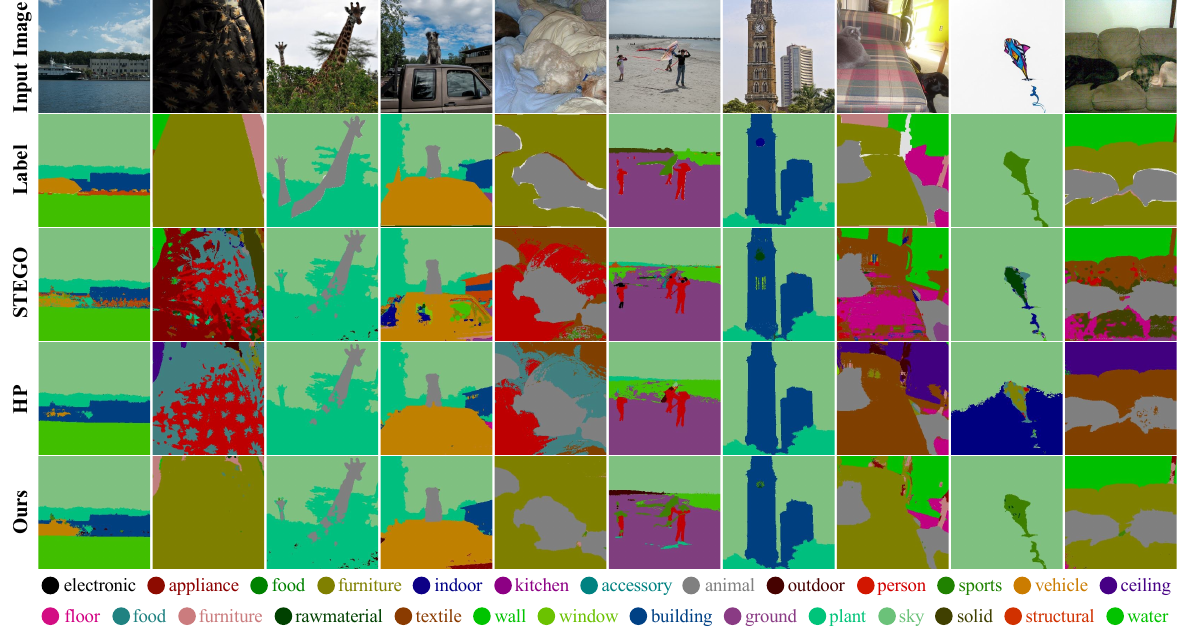}
  \caption{Additional qualitative results of COCO-Stuff dataset~\cite{coco} trained with ViT-S/8 backbone.
  }
  \label{fig:coco_supp}
\end{figure*}

\begin{figure*}[t!]
  \centering
  \includegraphics[width = .85\linewidth]{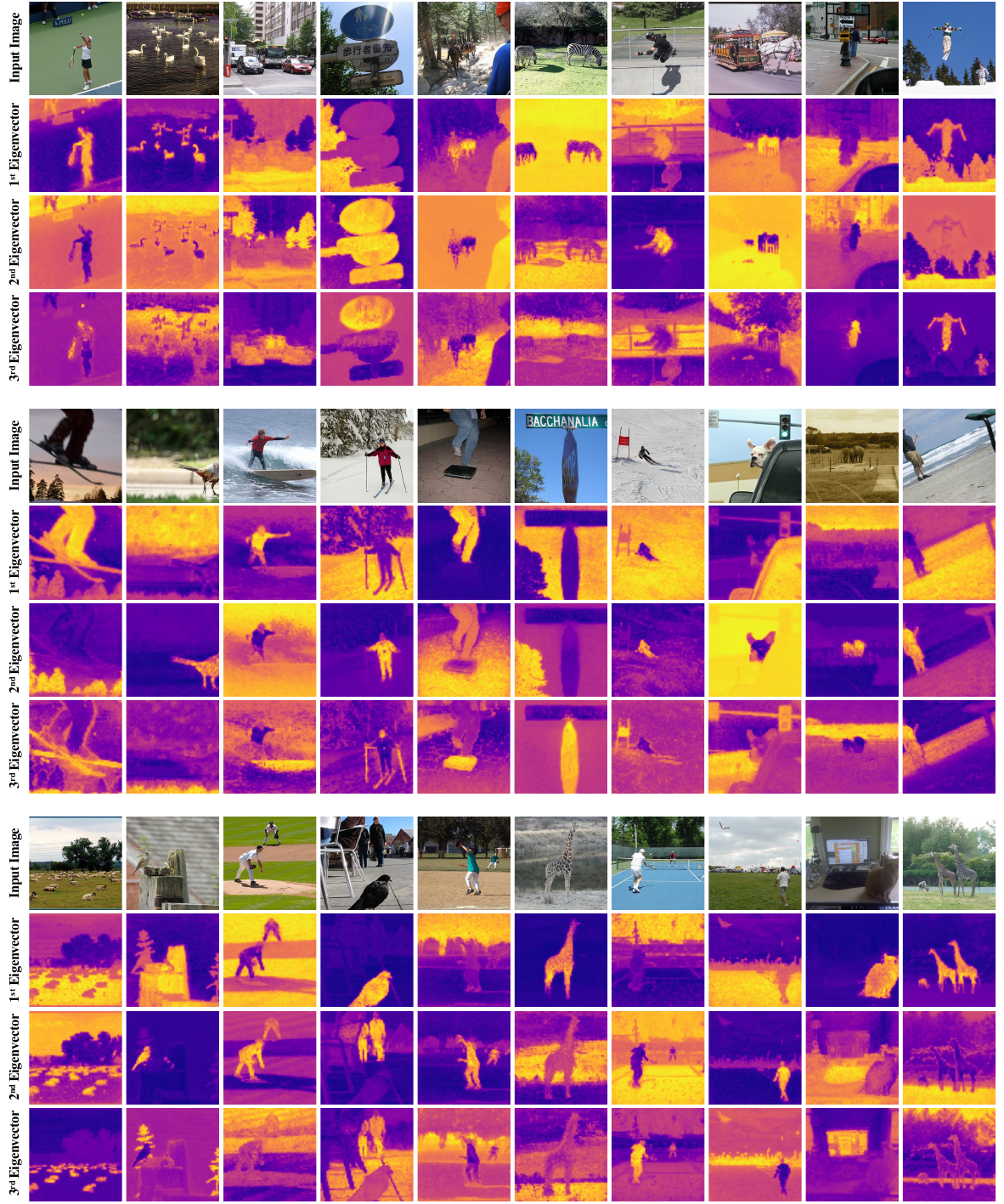}
  \caption{Additional visualization of the eigenvector obtained by training the COCO-Stuff dataset~\cite{coco} with ViT-S/8 as the backbone.
  }
  \label{fig:eigne_supp}
\end{figure*}

\begin{figure*}[t!]
  \centering
  \includegraphics[width = .85\linewidth]{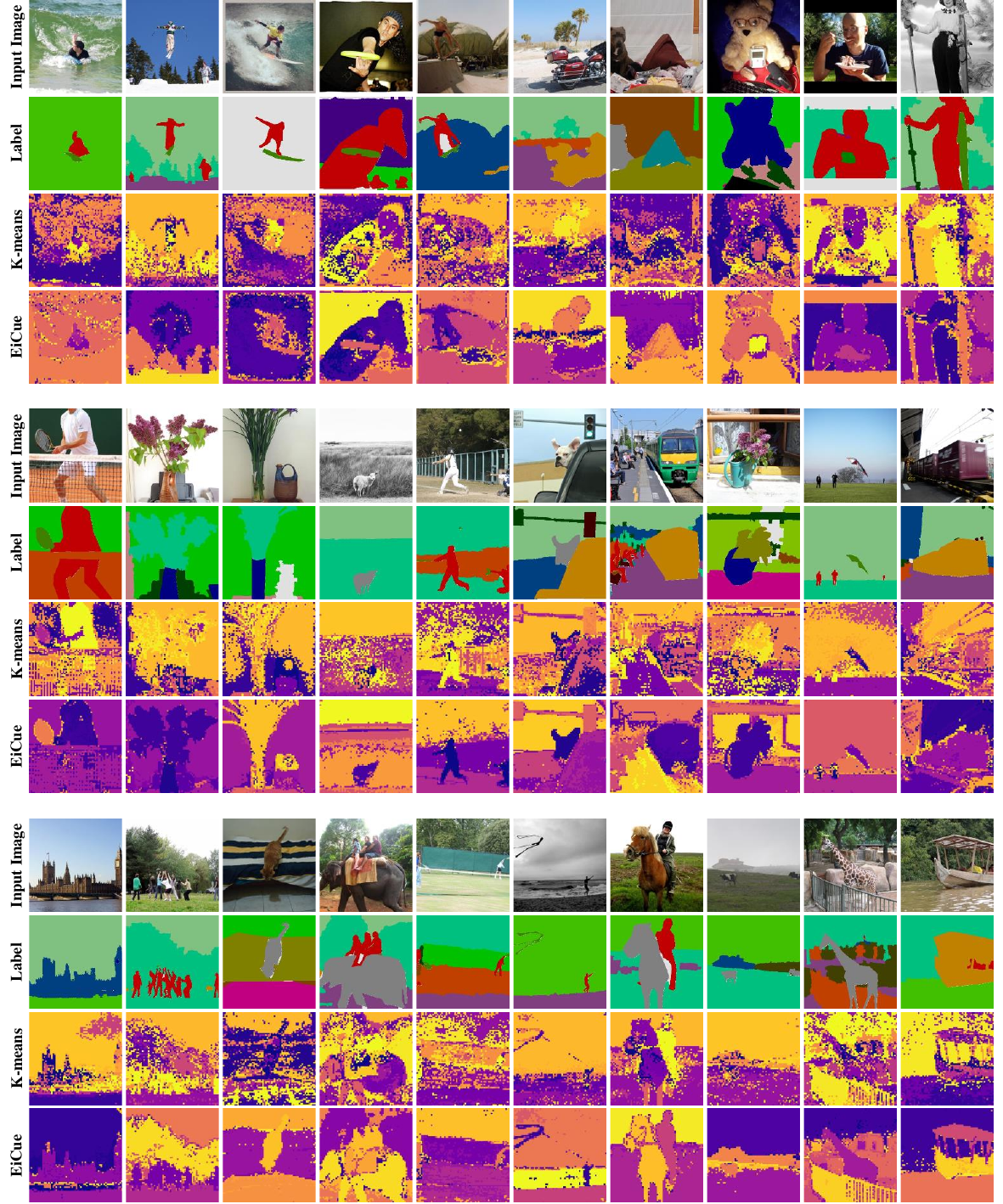}
  \caption{Additional comparison between K-means and our EiCue, with EiCue demonstrating enhanced performance in discerning object semantics and structures relative to K-means.
  }
  \label{fig:eicue_supp}
\end{figure*}

\end{document}